\documentclass{article}

\usepackage{microtype}
\usepackage{graphicx}
\usepackage{subfigure}
\usepackage{booktabs} % for professional tables
\usepackage{hyperref}

\usepackage[accepted]{icml2025}
\usepackage{etoolbox}

\usepackage{amsmath}
\usepackage{amssymb}
\usepackage{mathtools}
\usepackage{amsthm}

\usepackage[capitalize,noabbrev]{cleveref}

\theoremstyle{plain}

\theoremstyle{definition}

\theoremstyle{remark}

\usepackage[textsize=tiny]{todonotes}

\icmltitlerunning{ADT: Tuning Diffusion Models with Adversarial Supervision}

\begin{document}

\twocolumn[
\icmltitle{ADT: Tuning Diffusion Models with Adversarial Supervision}
\begin{icmlauthorlist}
\icmlauthor{Dazhong Shen}{SenseTime}
\icmlauthor{Guanglu Song}{SenseTime}
\icmlauthor{Yi Zhang}{SenseTime}
\icmlauthor{Bingqi Ma}{SenseTime}
\icmlauthor{Lujundong Li}{SenseTime}
\icmlauthor{Dongzhi Jiang}{SenseTime}
\icmlauthor{Zhuofan Zong}{SenseTime}
\icmlauthor{Yu Liu}{SenseTime}
\end{icmlauthorlist}
\icmlaffiliation{SenseTime}{SenseTime}

]




% \begin{abstract}
% This document provides a basic paper template and submission guidelines.
% Abstracts must be a single paragraph, ideally between 4--6 sentences long.
% Gross violations will trigger corrections at the camera-ready phase.
% \end{abstract}
\printAffiliationsAndNotice{}
\begin{abstract}
% Diffusion models are powerful for image generation by reversing the forward noising process to approximate true data distributions. 
% During training, models predict the clean conditional diffusion scores from the noised version of the true sample with only a single model pass, while inference involves iterative denoising steps from the white noise. 
% Diffusion models have achieved outstanding image generation by reversing a forward noising process to approximate true data distributions.
% While training involves the score prediction from the noised version of the true sample with a single model pass, inference requires iterative denoising from the white noise.
% Given the potential prediction biases, invertible error accumulation, and the various empirical inference tracks independent of the training process, such as classifier-free guidance, diffusion models are prone to suffer from training-inference divergences in practice, which limits the ability to align generative and real distribution.
%削减摘要
Diffusion models have achieved outstanding image generation by reversing a forward noising process to approximate true data distributions. 
During training, these models predict diffusion scores from noised versions of true samples in a single forward pass, while inference requires iterative denoising starting from white noise. 
This training-inference divergences hinder the alignment between inference and training data distributions, due to potential prediction biases and cumulative error accumulation.
% and the use of empirical inference techniques such as classifier-free guidance. 
% These factors hinder the alignment between generative and real data distributions. 
To address this problem, we propose an intuitive but effective fine-tuning framework, called Adversarial Diffusion Tuning (ADT), by stimulating the inference process during optimization and aligning the final outputs with training data by adversarial supervision. 
Specifically, to achieve robust adversarial training, ADT features a siamese-network discriminator with a fixed pre-trained backbone and lightweight trainable parameters, incorporates an image-to-image sampling strategy to smooth discriminative difficulties, and preserves the original diffusion loss to prevent discriminator hacking. 
In addition, we carefully constrain the backward-flowing path for back-propagating gradients along the inference path without incurring memory overload or gradient explosion. 
Finally, extensive experiments on Stable Diffusion models (v1.5, XL, and v3), demonstrate that ADT significantly improves both distribution alignment and image quality.

% Diffusion models excel in image generation by reversing a forward noising process to approximate true data distributions.
% While training involves the score prediction from the noised version of the true sample with a single model pass, inference requires iterative denoising from the white noise, leading to error accumulation and distribution misalignment due to the invertible prediction errors, especially with empirical techniques like classifier-free guidance, which are unavailable for the training process. 

% We propose the Adversarial Diffusion Model (ADM), which integrates the complete inference process into training through adversarial objectives. To ensure stable optimization, ADM introduces a siamese-network discriminator with a pre-trained backbone, employs dual-path sampling combining white noise and image-to-image generation, and maintains the original diffusion loss. We address the computational challenges of backpropagation through inference steps via selective gradient stopping, enabling efficient updates across all timesteps. Extensive experiments on state-of-the-art models like Stable Diffusion XL and SD3 demonstrate ADM's effectiveness in improving generation quality and distribution alignment.

\end{abstract}
\section{Introduction}
% background Diffusion Models DDPM GAN 
% In the past five years, diffusion models have proven to be a highly efficient and reliable framework for generative modeling. 
Diffusion models have recently been proven to be highly effective for image generation~\citep{sohl2015deep,ho2020denoising} with outstanding quality and generalization~\cite{podell2023sdxl,esser2024scaling}. 
In particular, 
the key feature of diffusion models is to approximate the training data distribution $p(x)$ by a reversal of a continuous Stochastic Differential Equation (SDE). 
The forward SDE progressively corrupts images $x$ with Gaussian noise $\epsilon_t$ as time grows, while the samples can be recovered by the reversed SDE with approximated score estimation $s_{\theta} \approx \nabla_x \log p(x,t)$ iteratively. 
% which directs the correct generation trajectory. 
% In particular, the backward SDE is theoretically linked to an Ordinary Differential Equation (ODE) with the same marginal distributions, offering an alternative way to generate samples in a deterministic  manner

% different from the previous generative models, such as Generative Adversarial Network (AGN)~\cite{}, and Variational Autoencoder (VAE)~\cite{}, where the whole inference process are stimulated in the optimization process
%VAE GAN 对生成链路的全过程进行优化
%介绍训练和推理的过程分别是什么样子
% Note that, different from the previous generative models, such as Generative Adversarial Network (AGN)~\cite{}, and Variational Autoencoder (VAE)~\cite{}, where the whole inference process is stimulated in the optimization process, diffusion models are prone to unavoidable errors arising from the training-inference discrepancy. Specifically, during each training step, the time step $t$ is first sampled from a continuous range or its densely discretized sequence. The diffusion model then takes the noised version $x_t$ of the ground truth samples $x_0$ as input. Only single step is performed to predict the current score $\nabla_x \log p(x,t)$.

\begin{figure}
    \centering
    \includegraphics[width=0.8\linewidth]{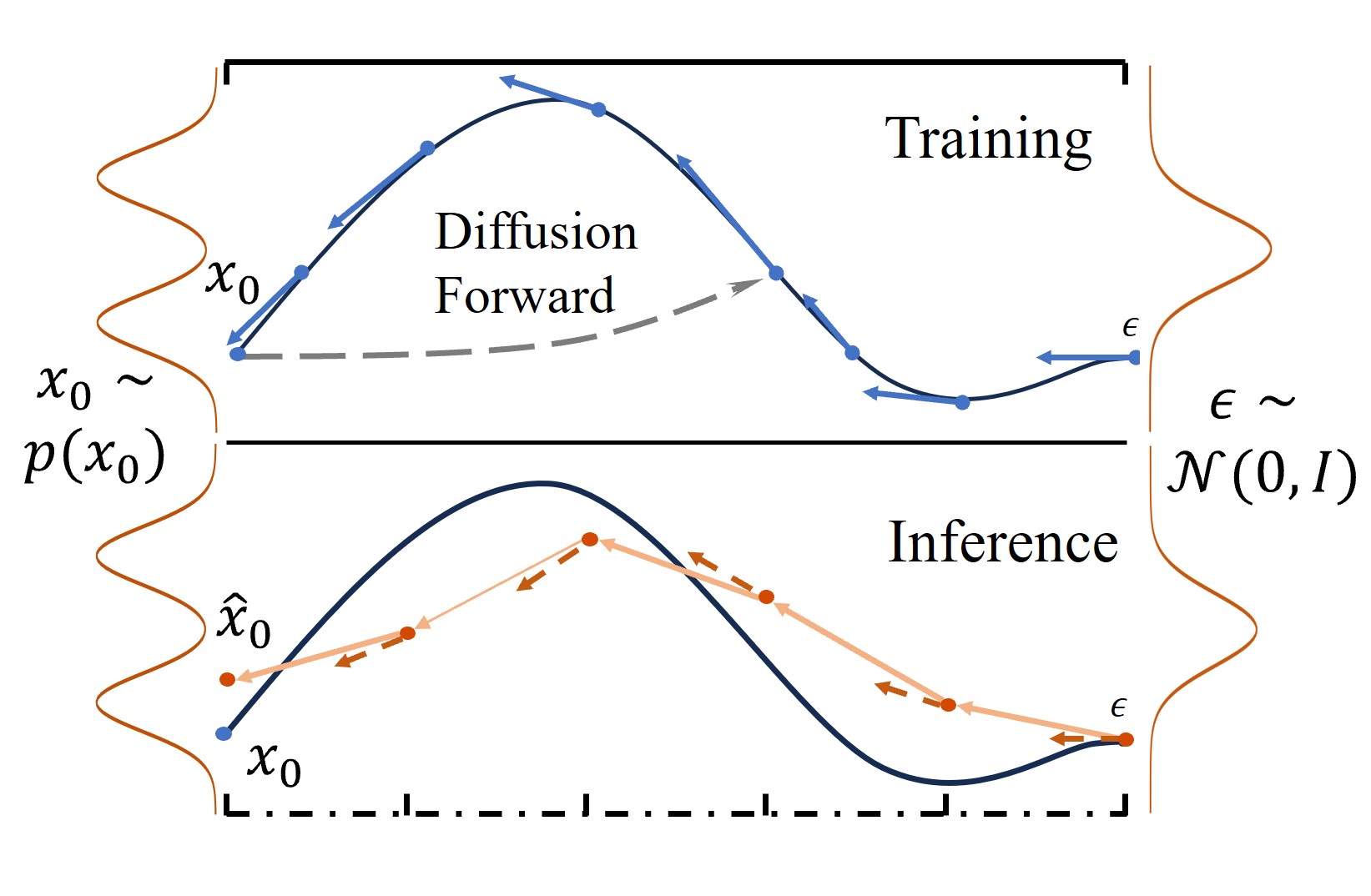}
    \vspace{-6mm}
    \caption{ A simple illustration for the training-inference divergence. The training procedure aims to fit the score function $\nabla_x \log p(x,t)$ at every time step with access to real data, while the inference turns to denoise from white noise through a fixed path without the exact score estimation and real data.
}
\vspace{-6mm}
    \label{fig:motivation}
\end{figure}

Unlike previous generative models such as Generative Adversarial Networks (GANs)~\citep{goodfellow2020generative} and Variational Autoencoders (VAEs)~\citep{kingma2013auto}, which integrate the entire inference process into the optimization, 
diffusion models are prone to suffer from training-inference discrepancies~\cite{ning2023input}. 
Specifically, as shown in Fig.~\ref{fig:motivation}, during each training step, a time step $t$ is first sampled randomly. 
% from a continuous range or its dense discretization. 
The diffusion model then predicts the current score $\nabla_x \log p(x,t)$ based on the noised version $x_t$ of the ground truth sample $x_0$, requiring only a single forward pass.
In contrast, during inference, samples are generated iteratively from white noise by denoising it through a fixed and sparse timestep sequence without access to the ground truth data. Each inference step is conducted based on the previously generated sample $\hat{x}_t$ and not exact score estimation $s_{\theta}$. In addition, conditional guidance strategies, such as classifier-free guidance~\citep{ho2022classifier}, may be employed in the text-to-image task, which is not available during training.
% crucial for text-to-image models
As a result, the training optimization cannot fully access the inference procedure and generative distribution in practice, limiting the ability for distribution alignment.

To align the generative and real data distribution, researchers have made several efforts to alleviate those training-inference divergences from different views. 
For instance, some works focus on aligning the model input errors between training and inference. They propose to perturb the training input~\citep{ning2023input} to stimulate the input bias during inference and enhance the model robustness, or scale model output~\citep{ningelucidating} and shift timesteps~\citep{lialleviating} to reduce sampling error accumulation along the inference path.
Other researchers try to modify the diffusion sampler to align the practical inference path and the trajectory of the original reversed SDE, such as deriving the analytic forms for different sampling function parts~\citep{bao2022analytic}, or applying different numerical solvers with varying accuracies and stability~\cite{nichol2021improved,karras2022elucidating}, such as DDIM~\cite{song2020denoising} and DPMSlover~\cite{lu2022dpm}.
However, given the various empirical inference tricks independent of the training process, such as using Classifier-free Gudiance~\cite{ho2022classifier} (CFG) or specifying inference timesteps, it is challenging to directly optimize the diverse errors and biases uniformly caused by training-inference divergences.
In this paper, we propose a unified fine-tuning framework called Adversarial Diffusion Tuning (ADT) to directly close the generative and training data distributions.
% we propose a new view to addressing the distribution dismatching cause by the training-inference divergences， and develop a unified fine-tuning framework called the Adversarial Diffusion Model (\textbf{ADM}) that addresses the training-inference divergences and matches the generative and real data distribution.
% without differentiating between individual gaps. 
Inspired by GANs and VAEs, our main idea is to simulate the inference process during optimization, allowing the final images to be affected by various practical errors, and using adversarial supervision to align them with training data.  
However, there exist two important challenges behind the optimization of ADT: 
1) \textbf{(Adversarial Optimization.)} Adversarial models are well-known for their unstable training~\cite{bai2021recent}, which can be exacerbated with the large-scale diffusion models.
To achieve robust optimization, we first propose a siamese-network-based discriminator with a frozen pre-trained image representation model as the backbone and a twin set of trainable lightweight discriminator heads, with the similarity-based discriminator score between the generative and real images. 
% The generative image is regarded as one augmented view of the real data, and the similarity of their head outputs is used as the discriminator score.
Additionally, instead of only inferring from the white noise, an image-to-image inference process, starting from a noisy image $ x_t $ at a random timestep $t$ is incorporated for smoothing the adversarial distance between the generative and real data.
Meanwhile, the original diffusion loss is preserved during optimization to prevent discriminator hacking.
% smoothing the adversarial distance through
% an image-to-image diffusion process
% This starting timestep can adjust the distance between the final generative image and the real image, making it easier to bridge their gap.
% Meanwhile, a loss weight schedule based on the starting timestep is designed to avoid excessive gradient changes when training early timesteps.
2) \textbf{(Backward-Flowing Path.)} The introduction of the iterative inference paradigm significantly increases memory consumption and the risk of gradient explosion when back-propagating gradients along the reversed inference path. 
Here,  we stop the gradients of the input for each iteration function except for one time of the input itself,
allowing the model at each inference step to be updated efficiently and equally, even for the very early step. 
% only sample a subset of timestep in the inference path for back-propagation. 
Finally, we validate the effectiveness of ADT on widely recognized text-to-image diffusion models with different model formulations, network architecture, and parameter scales, such as stable diffusion v1.5 (0.9B)~\citep{rombach2022high} and stable diffusion XL (3.5B)~\citep{podell2023sdxl} and stable diffusion 3 (2B)~\citep{esser2024scaling}.
Extensive experimental results show the advancements of ADT.

% 训练时间如何处理
% 如何处理Gradient Back-Propagating 
% SDXL-Turbo对比讨论
% DRtune 对比讨论    
% \input{sec/RelatedWorks}
% \vspace{-2mm}
\section{Preliminary}

\begin{figure*}[t]
    \centering
    \includegraphics[width=0.93\linewidth]{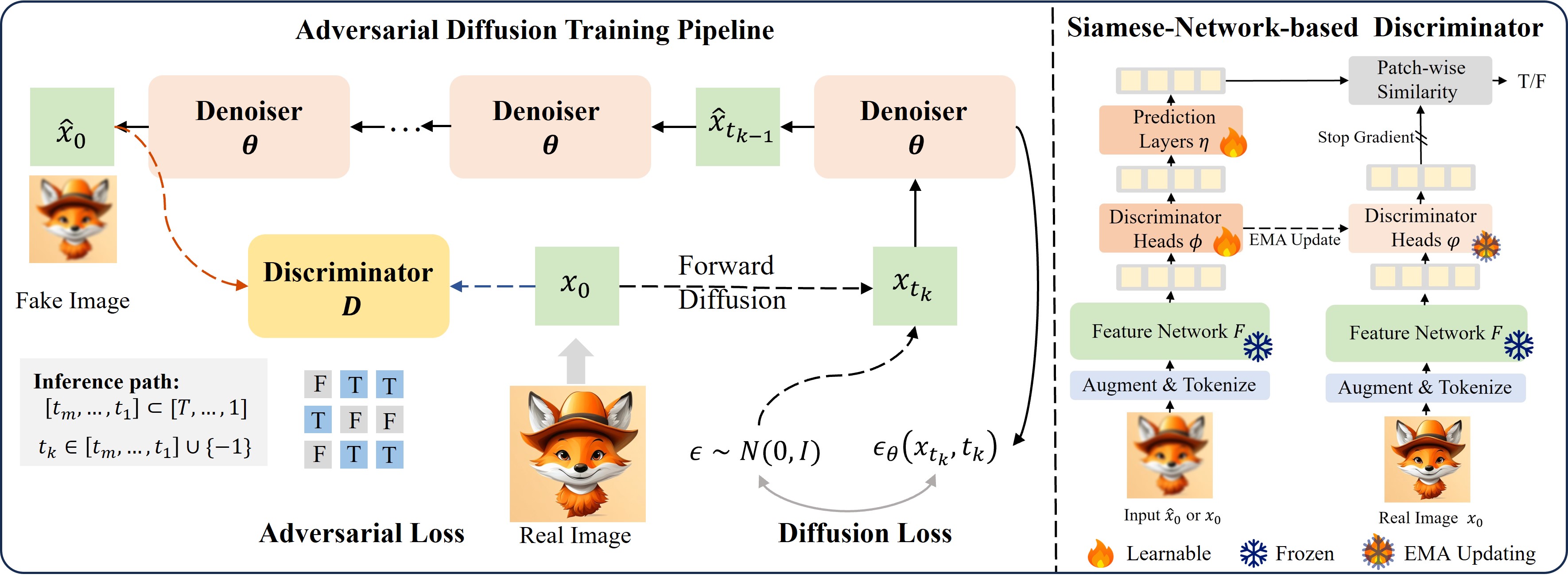}
    \vspace{-2mm}
    \caption{The overview framework of our ADT model. }
    \vspace{-4mm}
    \label{fig:framewok}
\end{figure*}

% \vspace{-2mm}
\subsection{Diffusion Models}
Given the image data space $\mathcal{X}$ with an unknown distribution $p(x_0)$. Diffusion models~\citep{sohl2015deep,ho2020denoising} define a forward process $\{x_t\}_{t\in [0,T]}$ for each image sample $x_0$ by a SDE equation:
\begin{equation}\label{equ:forwardSDE}
    \begin{split}
        {\rm d}  x_t = f(t) x_t {\rm d}  t + g(t){\rm d} w,~~x_0 \sim p(x_0),
    \end{split}
\end{equation}
where $w_t$ is the standard Wiener process, which indicates the image $x_0$ is corrupted by Gaussian noise as timestep $t$ grows. Specifically, for any $t$, the variable $x_t$ satisfies $x_t = \alpha_t x_0 + \sigma_t \epsilon, \epsilon\sim \mathcal{N}(0,I)$, where $\alpha_t$ and $\sigma_t$ are smooth scalar function of $t$ and referred as the noise schedule. In particular, when $t=T$, $x_T= \epsilon$.
% $q_T(x_T)\approx \mathcal{N} (0, \overline{\sigma}^2)$ for some $\overline{\sigma}^2>0$.
Under some regularity conditions, the reverse process of the above forward process has an associated probability flow ODE starting from time $T$, which has the same marginal distribution at time $0$ as that of the backward SDE~\citep{song2020score}, i,e.,
\begin{equation}\label{equ:backwardODE}
    \begin{split}
        {\rm d} x_t = \left [ f(t)  x_t - \frac{g^2(t)}{2}\nabla_x \log q_t(x_t)\right ] {\rm d} t. % ~x_T\sim q_T(x_T).
    \end{split}
\end{equation}
During learning, a neural network $\epsilon_{\theta}(x_t,t)$ would be used to estimate the scaled score function: $-\sigma_t\nabla_x \log q_t(x_t) $, which is also the predication of the used forward noise $\epsilon$. Therefore, the optimization objective is $\mathbb{E}_{x_0, \epsilon, t}[||\epsilon-\epsilon_{\theta}(x_t,t)||_2^2]$.
In practice, the inference process can be conducted by solving the ODE along a discrete timestep path $T=t_m>t_{m-1}>...>t_0=0$ with various ODE solvers, such as DDIM~\citep{song2020denoising}, DPMSlover~\citep{lu2022dpm}, etc. For all ODE solvers,  each inference step can be represented with an abstracted sampler:
\begin{equation}
    \begin{split}
        x_{t_{i-1}} = a_{t_i} x_{t_i} +b_{t_i}\epsilon_{\theta}(x_{t_i}, t_i),
    \end{split}
\end{equation}
where $a_{t_i}$ and $b_{t_i}$ can be determined by noise schedules and vary depending on the ODE solver used.
In addition, for the text-to-image task, the Classifier-Free Guidance (CFG)~\cite{ho2022classifier} strategy is widely used to follow the conditional prompt $c$ with a conditional noise predictor $\epsilon_{\theta}(x_t,c,t)$. Specifically, CFG combines the unconditional/conditional models and obtains a new noise predictor: $\epsilon^{'}_{\theta}(x_{t_i},c,t_i) = (\gamma-1)\epsilon_{\theta}(x_{t_i},t_i) + \gamma \epsilon_{\theta}(x_{t_i},c,t_i)$, where $\gamma$ is the guidance scale that controls the image-prompt alignment. 
Note that, for simplicity, we only use the signals from unconditional sampling. This approach can be easily extended to scenarios involving the CFG strategy
% Note that, for simpility, we only use the signals of unconditional sampling, which can be easily extended for the scenario wit the CFG strategy.
% Without the lack of generalization, we begin the following discussion based only on unconditional sampling, which can be extended to the text-to-image task with the CFG strategy.

% Specifically,  CFG randomly takes the prompt $c$ as another input of the network $\epsilon_{\theta}$ to model the conditional diffusion score, i.e., $\epsilon_{\theta}(x_t,c,t) \approx  -\sigma_t\nabla_x\log q_t(x_t|c)$.  After optimization,
% CFG replaces the unconditional noise estimation $\epsilon_{\theta}(x_{t_i}, t_i)$ during sampling with a new predictor, i.e., 
% \begin{equation}\label{equ:cfg}
%     \begin{split}
%         \epsilon^{'}_{\theta}(x_{t_i},c,t_i) = \epsilon_{\theta}(x_{t_i},t_i) + \gamma (\epsilon_{\theta}(x_{t_i},c,t_i) -\epsilon_{\theta}(x_{t_i},t_i)),
%     \end{split}
% \end{equation}
% where the second item on the right estimates the classifier's gradients $\nabla_x \log p(c|x_t)$, and $\gamma$ is the guidance scale that controls the image-prompt alignment.

 %要不要解说三种gaps产生的原因
\subsection{Back-Propagation for Iterative Sampling}
Optimizing diffusion models using gradients on the output image from iterative inference has become crucial for capturing properties that are challenging to address in a single diffusion step, such as human preference~\citep{kirstain2023pick,wu2023human} and aesthetic evaluation~\citep{murray2012ava}. However,  a dilemma will arise: back-propagating gradients through the inference path to earlier steps incurs substantial computational overhead and risks gradient explosion~\citep{clark2023directly}. To mitigate these risks, some works~\citep{clark2023directly,xu2024imagereward,prabhudesai2023aligning} constrain the back-propagation path to inference steps near the output image. Nevertheless, training only on the last few steps may be insufficient.
More recently, DRTune~\citep{wu2024deep} proposes blocking the gradient of the input for each noise estimation \(\epsilon_{\theta}\), i.e.,
\begin{equation}\label{equ:DRTune}
    \begin{split}
        x_{t_{i-1}} = a_{t_i} x_{t_i} +b_{t_i}\epsilon_{\theta}(sg(x_{t_i}), t_i),
    \end{split}
\end{equation}
where $sg(\cdot)$ denotes the stop gradient operation, allowing the gradient to flow only through the term $a_{t_i}x_{t_i}$. In this manner, each inference step can be independently optimized, 
% using $\partial x_0/\partial x_{t_i} = \prod_{s=1}^t a_{t_i}$
providing a stable way to control the backward gradient flow along the entire inference path. To further reduce memory, only a subset of inference steps can be sampled for training.
Here, we further extend DRTune for the gradient back-propagation from the adversarial loss on the output image for stable and effective training.

\vspace{-2mm}
\section{Adversarial Diffusion Tuning}
\vspace{-1mm}
% In this section, we present the technical details of our ADM model, which aims to align the distribution of the final inference image of diffusion models with that of the real data. 
Our core idea is to incorporate the entire inference process into the model training process, where the inference images would be accessible for model optimization and are optimized in an adversarial manner.
% to sample from the manifold of inference images. 
% Given the absence of closed-form expressions for the distributions of both real and inference images, the adversarial objective effectively measures the distance between them, requiring only image sampling.
The left part of Figure~\ref{fig:framewok} shows the overview training pipeline of ADT which involves two learnable networks: the pretrained diffusion model $\epsilon_{\theta}$ with weights $\theta$ as the generator $G_{\theta}$, and a discriminator $D_{\phi}$ with trainable weights $\phi$.
During training, the same sampling strategy $\pi$ as that used during inference is applied to generate samples $\hat{x}_0$ and stimulate the generative image space  $\mathcal{X}^{\pi}_{G_{\theta}}$. With ture samples $x_0\in \mathcal{X}$, the training objective is to optimize the  adversarial objective:
\begin{equation}\label{equ:advloss1}
    \begin{split}
        \min_{\theta} \max_{\phi} \mathcal{L}_{\rm adv} (\mathcal{X}, \mathcal{X}^{\pi}_{G_{\theta}} ; G_{\theta}, D_{\phi}, \pi)= \\
        \mathbb{E}_{x_0 \in \mathcal{X}}[\log (D_{\phi}(x_0))] + \mathbb{E}_{\hat{x}_0 \in \mathcal{X}_{G_{\theta}}^{\pi} }[\log (1- D_{\phi}(\hat{x}_0))].
    \end{split}
\end{equation}

Specifically, to ensure \textbf{training stability} during adversarial optimization, we enhance the overall procedure in three key aspects:  reducing the number of learnable parameters in the discriminator, smoothing the adversarial distance through an image-to-image diffusion process, and incorporating the original diffusion loss to prevent knowledge forgetting. 
Additionally, for \textbf{efficient gradient back-propagation} through the inference path, we follow~\citep{wu2023human} and use the stop gradient operation to constrain and deepen the backward-flowing path, allowing the model at each inference step to be updated equally without memory overload or gradient explosion.
The overall training process can be found in Algorithm~\ref{alg:training}.

\subsection{Adversarial Optimization}
\vspace{-1mm}
\textbf{Siamese-Network based Discriminator.} For the discriminator $D$, we follow~\citep{sauer2023stylegan} and use a frozen pre-trained feature network $F$ with some trainable lightweight discriminator heads $h_1, \ldots, h_m$. The right part of Figure~\ref{fig:framewok} shows the overview network architecture. 
Specifically, We utilize DINOv2~\citep{oquab2023dinov2} as the feature network $F$, leveraging its state-of-the-art self-supervised visual representations learned from large-scale datasets.
% DINO V2~\citep{oquab2023dinov2}is used as the feature network $F$ which achieve the start-of-the-art visual representation capability based on self-supervised learning on large-scale datasets.
% Then, with ViT~\citep{dosovitskiy2020image} as the DINO V2's backbone, where each input image $x_I = \hat{x}_0$ or $x_0$ is split into a patch token sequence, we define small convolution layers as the discrimination heads $h_i$ on the patch embeddings extracted from certain transformer blocks. 
DINOv2 is implemented with a ViT backbone~\citep{dosovitskiy2020image}. Each input image $x_I = \hat{x}_0$ or $x_0$ is first tokenized into a sequence of patches. We apply small convolutional discrimination heads $h_i$ to the patch embeddings extracted from selected transformer blocks.

% \begin{figure}
%     \centering
%     \includegraphics[width=1.0\linewidth]{figs/Discriminator.jpg}
%     \caption{The network architecture of the siamese-network based discriminator. }
%     \label{fig:discriminator}
% \end{figure}

In particular, instead of directly outputting a scalar score by the discriminator head $h_i$, we introduce a dual network $h_i^{'}$ for each $h_i$ with different weights $\varphi$. The discrimination score is defined as the similarity between their output vectors $z_I$ and $z_R$, corresponding to the input image $x_I$ and the reference real image $x_0$ with the same prompt $c$, respectively.
% we induce the siamese network $h_i^{'}$ for each $h_i$ where the feature embeddings of the image $x_I$ and the reference real image $x_0$ with the same prompt $c$ are their inputs, respectively, and the outputs of both them are vectors.
% Finally, the patch-wise similarity of the head outputs from input and reference images is regarded as the discriminator score. 
The idea behind this is inspired by the self-supervised learning  approaches~\citep{grill2020bootstrap,chen2020simple}, 
% where two siamese networks with the same architecture embed different augmented views of the same image with similar outputs.  
where the generative image $\hat{x}_0$ is regarded as the augmented view of the corresponding reference image $x_0$. 
The generator is optimized to generate similar images to reference images so that the corresponding output vectors $z_I$ and $z_R$ are close to each other.
% that can be embedded as a similar output to the reference image by the discriminator.
More specifically, we follow~\citep{grill2020bootstrap} and introduce a simple predictor network $g_{\eta}$ on the vector $z_I$ to predict the vector $z_R$. The discriminator score is defined as,
\begin{equation}
    \begin{split}
        D_{i} (x_I) =  \frac{<q_{\eta}(h_i(F(x_I))), h^{'}_i(F(x_0))>}{||q_{\eta}(h_i(F(x_I)))||_2\cdot |h^{'}_i(F(x_0))||_2}.
    \end{split}
\end{equation}
where, the parameters $\varphi$ in the dual head $h_i^{'}$ is without gradient backward and updated in each training step by the exponential moving average of the parameters $\phi$ of the corresponding head $h_i$ , i.e., $\varphi \leftarrow \tau \varphi + (1-\tau) \phi$ with the decay rate $\tau\in [0,1]$. 
% Therefore, the discriminator is optimized by the following  equations:
% \begin{equation}
%     \begin{split}
%         \min_{\phi, \eta} \mathbb{E}_{x_0 \sim \mathcal{X}}[\log (D_{\phi,\eta}(x_0))] + \mathbb{E}_{\hat{x}_0 \in \mathcal{X}_{G_{\theta}}^{\pi} }[\log (1- D_{\phi,\eta}(\hat{x}_0))]\\
%         \varphi \leftarrow \gamma \varphi + (1-\gamma) \phi
%     \end{split}
% \end{equation}
% {\color{red} more details for ema}

Previous works report that two Siamese networks benefit from a robust optimization process~\cite{he2020momentum,grill2020bootstrap}, where we have also observed better performance in our practice. 
In particular, we apply differentiable data augmentation~\citep{zhao2020differentiable} to the input image, $\hat{x}_0$ or $x_0$, which has been proven to significantly improve performance~\citep{sauer2023stylegan}.

\begin{algorithm}[t]
\caption{The Training Procedure of ADM}\label{alg:training}

\begin{algorithmic}[1]
    \STATE \textbf{Input:}  training set $X$, generator $G_{\theta}$ with pre-trained model $\epsilon_{\theta}$, discriminator $D$ with parameters $\{\phi, \varphi, \eta \}$, sampling strategy $\pi$ with the timestep set $\{t_m, \ldots, t_0\}$, learning rate $\gamma$, EMA rate $\tau$, hyper-parameters $\lambda$, $H$, $K$
    % \STATE \textbf{Output:} Output description
    % \STATE iter = 1
    % \WHILE{ not convergence }
    \FOR{iter $=1,2,..., N$} 
    \STATE Sample $x_0\in X$,  $s \in [m,\ldots, 0]$, $\epsilon \sim \mathcal{N}(0, I)$
    % \STATE \textbf{if} {$t_s=-1$} \textbf{then} $x_{t_s} =  \epsilon$  \textbf{else}  $x_{t_s} = \alpha_{t_s}x_0 + \sigma_{t_s} \epsilon$ 
    % \STATE Infer sample $\hat{x}_0 =\mathcal{T}^{\pi}_{t_s\rightarrow 0}(x_{t_s}; G_{\theta}) $
    % \IF{$t_s=-1$}
    %     \STATE Smaple $\hat{x}_0 = \pi(\epsilon; G_{\theta})$ and output $\epsilon_{\theta}(\epsilon, T)$%$x_{t_s} = \epsilon$
    % \ELSE
    % \STATE Sample $\hat{x}_0 =\mathcal{T}^{\pi}_{t_s\rightarrow 0}(x_0, \epsilon; G_{\theta}) $ 
    % \STATE  $\epsilon_{\theta} = \epsilon_{\theta}(x_{t_s}, t_s)$
    \STATE $\hat{x}_0=\alpha_{t_s}x_0 + \beta_{t_s} \epsilon$, $k = \min(K, s)$
    \STATE Sample $\mathbb{S} = \{t_{k_j}\}_{j=1}^k,  \subset \{t_s, \ldots, t_1\}$
    \FOR{$t = t_s\ldots, t_1$}
        \IF{$t=t_s$}
            \STATE  $\overline{\epsilon}=\hat{\epsilon}=\epsilon_{\theta}(x, t)$
        \ELSIF{$t \in \mathbb{S}$}
            \STATE $\hat{\epsilon}=\epsilon_{\theta}(sg(x), t)$
            % \STATE \textbf{if} {$t=t_s$} \textbf{then} $\hat{\epsilon}_{t_s} = \hat{\epsilon}$ 
        \ELSE
            \STATE $\hat{\epsilon}=sg(\epsilon_{\theta}(x, t))$
        \ENDIF

        \STATE $\hat{x}_0 =\hat{x}_0+(a_{t}-1) sg(\hat{x}_0) +b_{t}\hat{\epsilon}$
    \ENDFOR
    
    % \ENDIF
    \STATE Compute $\mathcal{L}^D_{adv} $  in Equ.~\ref{equ:advlossD} with $x_0$ and $\hat{x}_0$
    \STATE Update parameter $[\phi, \eta] \leftarrow  [\phi, \eta] -\beta \nabla_{\phi, \eta} \mathcal{L}^D_{adv}$,
    \STATE Update parameter $\varphi \leftarrow \tau \varphi +(1-\tau) \phi$
    \IF{iter$\% H=0$}
        \STATE Compute $\mathcal{L}^G $ in Equ.~\ref{equ:advlossG} with $\hat{x}_0$, $\overline{\epsilon}$ and $\lambda$
        \STATE Update generator $\theta\leftarrow \theta - \gamma \nabla_{\theta} \mathcal{L}^G$ 
    \ENDIF
    % \STATE iter+=1
    % \ENDWHILE
    \ENDFOR
\end{algorithmic}
\end{algorithm}

\noindent\textbf{Image-to-Image Inference.}
% To alleviate the errors from various training-inference divergences, simulating the whole inference process during optimization is necessary, where the paradigm actually relies solely on prompts and starts inference from the white noise~\citep{}. 
% To alleviate errors arising from various training-inference divergences, it is necessary to simulate the entire inference process during optimization, which actually relies solely on prompts and initiates inference from white noise~\cite{}. 
% However, due to these errors, there is a significant gap between the generative image distribution and the real data distribution. 
Actually, in the early training stages of ADT, there is a significant gap between the generative distribution and training data distribution, which can be recognized by the discriminator easily but is challenging for the generator to close.
% This gap is easily recognized by the discriminator but challenging for the generator to close.
As a result, the training process would be unstable due to the optimization imbalance between the generator and discriminator.
To address this problem, we further simulate the image-to-image inference process into the optimization. 

Specifically, during each training iteration, the starting timestep $t_s$ is first uniformly selected from the entire discrete inference path $\{T, t_{m-1}, \ldots, 0\}$, and the initial latent image is defined as $x_{t_s} = \alpha_{t_s} x_0 + \sigma_{t_s} \epsilon$, where the noise $\epsilon$ is drawn from $\mathcal{N}(0, I)$.
Intuitively, when $t_s$ approaches 0, the initial image more closely resembles $x_0$, resulting in an output image $\hat{x}_0$ that better aligns with the real sample $x_0$. This provides a natural mechanism to reduce the divergence between generated outputs and training data.
% Intuitively, when $t_s$ is close to $0$, the initial image is more like $x_0 $. Correspondingly, the inference image $\hat{x}_0$ would resemble the real sample $x_0$ more, which provides a way to smooth the divergence between the outputs and real data.
% whereas a larger difference occurs when $t_s$ is near $T$. 
Formally, denoting $\mathcal{T}^{\pi}_{t_s\rightarrow 0}$ as the image-to-image inference path starting from timestep $t_s$ along with the sampling strategy $\pi$.
% \begin{equation}
%     \begin{split}
%         KL(p(x_0;\mathcal{T}^{\pi}_{t_s\rightarrow 0}) || q_0(x_0)) >  KL(p(x_0;\mathcal{T}^{\pi}_{t_{s'}\rightarrow 0}) || q_0(x_0))
%     \end{split}
% \end{equation}
% where $t_s>t_{s'} \in \{t_m, \ldots, t_0\}$
The adversarial loss in Equation~\ref{equ:advloss1} would be extended as:
\begin{equation}\label{equ:advloss2}
    \begin{split}
        \mathcal{L}_{\rm adv}= \mathbb{E}_{t_s\sim \mathcal{U} (\pi)} [\mathcal{L}_{\rm adv} (\mathcal{X}, \mathcal{X}_{G_{\theta}} ; G_{\theta}, D_{\phi}, \mathcal{T}^{\pi}_{t_s\rightarrow 0})].
    \end{split}
\end{equation}
% where we define $\mathcal{T}^{\pi}_{t_s\rightarrow 0} = \pi$ when $t_s=-1$, which indicates that the inference path starts from the white noise. 
In particular, the CFG strategy is incorporated into each inference path $\mathcal{T}^{\pi}_{t_s \rightarrow 0}$ to align with practical inferences.

\noindent\textbf{Joint Learning with Diffusion Loss.}
% Here, we summarize the whole training procedure as Algorithm~\ref{alg:traing}. 
% Given the pre-trained diffusion model $\epsilon_{\theta}$ as the generator $G$, the discriminator $D$ defined in Figure~\ref{}, and the sampling strategy $\pi$, we follow~\citep{sauer2023stylegan,gulrajani2017improved}, and train the discriminator and generator alternately based on adversarial loss in Equation~\ref{equ:advloss1}. 
Note that, the optimizing direction of our adversarial loss may be largely different from that of the original diffusion loss, which may cause the problem of catastrophic forgetting, even the collapse of image quality. 
% The underlying reason can be that the optimized generative data distribution $\mathcal{X}^{'}_{G}$ may be distant from the initialization one $\mathcal{X}_G$,
% which increases the possibility for the generator to hack the discriminator.
% This may occur because the optimized generative data distribution $\mathcal{X}^{'}_{G}$ diverges from the initial distribution $\mathcal{X}_G$, increasing the likelihood of the generator hacking the discriminator.
Therefore, we also preserve the original diffusion loss $\mathcal{L}_{diff}$ to guide the updating for the generator $\epsilon_{\theta}$. In addition, we use hinge loss~\citep{lim2017geometric} as the practical adversarial objective function. Then,  the generator's objective amounts to:
\begin{equation}\label{equ:advlossG}
    \begin{split}
        \mathcal{L}^{G} = \mathbb{E}_{\tiny \hspace{-1mm}\begin{array}{l}
              x_0\in\mathcal{X}\\
               t_s\sim \mathcal{U} (\pi) \\
               \epsilon \sim \mathcal{N}(0,I)
        \end{array}\hspace{-1mm}}  [-\sum_i D_{i}(\mathcal{T}^{\pi}_{t_s\rightarrow 0}(x_0, \epsilon; G_{\theta})) \\  + \lambda ||\epsilon_{\theta}(x_{t_s},{t_s})-\epsilon||^2_2 ], 
    \end{split}
\end{equation}
% \begin{equation}\label{equ:advlossG}
%     \begin{split}
%         \mathcal{L}^{G} = \mathbb{E}_{\tiny \hspace{-1mm}\begin{array}{l}
%               x_0\in\mathcal{X}\\
%                t_s\sim \mathcal{U} (\pi) \\
%                \epsilon \sim \mathcal{N}(0,I)
%         \end{array}\hspace{-1mm}} \left [\underbrace{-\sum_i D_{i}(\mathcal{T}^{\pi}_{t_s\rightarrow 0}(x_0, \epsilon; G_{\theta})) }_{\mathcal{L}^G_{adv}} \right . \\
%         \left. + \lambda \underbrace{||\epsilon_{\theta}(x_{t_s},{t_s})-\epsilon||^2_2}_{\mathcal{L}_{diff}}\right ], 
%     \end{split}
% \end{equation}
where the first line represents the generator loss indicated as $\mathcal{L}_{adv}^G$ and the second line specifies the diffusion loss $\mathcal{L}_{diff}$, applied at the start timestep $t_s$ of the inference path.
$\lambda$ is a hyper-parameter to balance the above two loss items.
% In particular, $\mathcal{L}_{diff}$ is applied at the start timestep $t_s$ of the inference path, avoiding additional complexity for computing diffusion loss.
Correspondingly, the discriminator $D$ is updated by minimizing the following objective,
\begin{equation}\label{equ:advlossD}
    \begin{split}
        \mathcal{L}_{\rm adv}^{D} = \mathbb{E}_{\tiny \hspace{-1mm}\begin{array}{l}
              x_0\in \mathcal{X}\\
               t_s\sim \mathcal{U} (\pi) \\
               \epsilon \sim \mathcal{N}(0,I)
        \end{array}\hspace{-1mm}}  [\sum_i {\rm ReLU}( 1-D_{i}(x_0))  \\
         + {\rm ReLU}( 1+ D_{i}(\mathcal{T}^{\pi}_{t_s\rightarrow 0}(x_0, \epsilon; G_{\theta})))  ], 
    \end{split}
\end{equation}
In addition, to further balance the optimization of the generator and discriminator, we reduce the updating frequency of the generator to $1/H$, since the generator tends to easily hack the discriminator in our practice.

% with different frequencies. 
% Intuitively, the 

%Start from noise VS image-to-image
%流程描述
%MSE loss 
% DDIM 50 CFG 6.0

%DINO
%patch-wise Similarity +CLIP 
% 
% \begin{figure}
%     \centering
%     \includegraphics[width=0.4\linewidth]{figs/GradScale.jpg}
%     \caption{The dynamic curve of gradient scale with different noise schedules.}
%     % where the early inference step (near 1000) suffers from the gradient explosion
%     \vspace{-2mm}
%     \label{fig:gradscale}
% \end{figure}

% \subsection{Gradient Flow Constraint Mechanism}
\subsection{Backward-Flowing Path Constraint}
\vspace{-1mm}
% Improve DRTune 
%给出相关证明 表达出 优势
% memory consumption
% depth-efficiency 
At each training iteration for the diffusion model $\epsilon_{\theta}$, we need to back-propagate the gradient step-by-step from the final output image $\hat{x}_0$ to latent image $x_{t_s}$ at the starting timestep $t_s$, where the memory consumption limit the depth of gradient back-propagation.
A simple solution is DRTune~\citep{wu2024deep}, which applies the stop gradient operation on the model input during forward inference, as shown in Equ.~\ref{equ:DRTune}. However, we argue that \textit{DRTune suffers from rapid gradient explosion when backpropagating to the early inference timesteps}.
Specifically, for the loss function $\mathcal{L}$ on the final output $\hat{x}_0$, the partial differential derivative on the latent image $\hat{x}_{t_i}$ can be derived:
\begin{equation} \label{equ:dloss_dxt}
    \begin{split}
        \frac{\partial \mathcal{L}(\hat{x}_0)}{\partial \hat{x}_{t_i}} = \frac{\partial  \mathcal{L}(\hat{x}_0)}{\partial \hat{x}_0} \frac{\partial \hat{x}_0}{\partial \hat{x}_{t_i}}= \frac{\partial  \mathcal{L}(\hat{x}_0)}{\partial \hat{x}_0} \prod_{j=1}^i a_{t_j},
    \end{split}
\end{equation}
where $a_{t_i}$ increases to a very large size when $t_i$ grows to $T$ with the common used linear or cosine noise scheduler~\cite{chen2023importance} (See Appendix for more discussions). 
% As the example in Figure~\ref{fig:gradscale}, 
As a result, the gradient scale will explode rapidly for the early inference steps and lead to unstable optimization, where the scenario becomes worse in our adversarial training. 
% In addition, considering every single sampling step contributes to the final output, overly unbalanced weights among backward gradients for different timesteps cause the model to overlook the final inference steps, which complete the image details and improve quality. Instead, it focuses on the early inference steps, only constructing the semantic structure of the full image.

% As a result, similar to previous works~\citep{}, DRTune tends to overlook the contributions of early inference steps, which are crucial for constructing the semantic structure of the full image. Instead, it focuses on the final inference steps, which primarily complete the image details.
% can neither deliver the gradient to the early inference timestep 

To balance the gradient scale on inference steps, we further propose to add the stop gradient operation on the linear part of the sampling equation, i.e., 
\begin{equation}
\small
    \begin{split}
        x_{t_{i-1}} =x_{t_i}+(a_{t_i}-1) sg(x_{t_i}) +b_{t_i}\epsilon_{\theta}(sg(x_{t_i}), t_i),
    \end{split}
\end{equation}
Then, the backward gradient flows through the inference path without scale variation, i.e., ${\partial L(\hat{x}_0)}/{\partial \hat{x}_{t_i}} = {\partial L(\hat{x}_0)}/{\partial \hat{x}_0}$. Similar to DRTune~\cite{wu2024deep}, we select only $K$ inference steps $\mathbb{S} = \{t_{k_j}\}_{j=1}^K$ for training; otherwise, the model $\epsilon_{\theta}$ becomes detached from the backward-flowing path. The detailed forward inference process during training is presented at line 4-15 in Alg.~\ref{alg:training}.
The corresponding updating gradient can be derived as:
\begin{equation}
\small
    \begin{split}
        \nabla_{\theta} \mathcal{L}^G = \frac{\partial \mathcal{L}^G_{adv}(\hat{x}_0)}{\partial \hat{x}_0} \sum_{t_i \in \mathbb{S}} b_{t_i} \frac{\partial \epsilon_{\theta}(\hat{x}_{t_i}, t_i)}{\partial \theta} + \lambda \frac{\partial \mathcal{L}_{diff}}{\partial \theta }.
    \end{split}
\end{equation}

% \begin{algorithm}[t]
% \centering
% \caption{Gradient Flow Constraint Mechanism}\label{alg:gradflow}
% \begin{algorithmic}[1]
%     \INPUT: diffusion model $\epsilon_{\theta}$, sampling strategy $\pi$ with the timestep set $\{t_m, \ldots, t_0\}$, number of training timesteps $K$, starting timestep $t_s$ with latent image $x_{t_s}$

%     \STATE $x=x_{t_s}$, $K = \min(K, s)$
%     \STATE Sample $\mathbb{S} = \{t_{k_j}\}_{j=1}^K \subset \{t_s, \ldots, t_1\}$
%     \FOR{$t = t_s\ldots, t_1$}
%         \IF{$t \in \mathbb{S}$}
%             \STATE $\hat{\epsilon}=\epsilon_{\theta}(sg(x), t)$
%             % \STATE \textbf{if} {$t=t_s$} \textbf{then} $\hat{\epsilon}_{t_s} = \hat{\epsilon}$ 
%         \ELSE
%             \STATE $\hat{\epsilon}=sg(\epsilon_{\theta}(x, t))$
%         \ENDIF

%         \STATE $x =x+(a_{t}-1) sg(x) +b_{t}\hat{\epsilon}$
%     \ENDFOR
%     \OUTPUT: $x$  %and $\hat{\epsilon}_{t_s}$
% \end{algorithmic}
% \end{algorithm}
% \vspace{-4mm}

\subsection{Extension to Flow-Matching Model}
\vspace{-1mm}
In the above discussion, we take DDPM as the representative model. Actually, ADT can be easily extended to the flow-matching model~\cite {liu2022flow,esser2024scaling}, which is another state-of-the-art variant of diffusion models and has also been used widely~\cite{esser2024scaling,flux}.
Flow-matching models simplify the forward diffusion process as the linear interpolation of the real data and the white noise $\epsilon\sim\mathcal{N}(0,I)$, i.e., $x_t = (1-t)x_0 +t\epsilon, t\in [0,1]$. A neural network $v_{\theta}(x_t, t)$ is optimized to estimate the conditional velocity $v(\epsilon)=\epsilon-x_0$ with the diffusion optimization loss $\mathcal{L}_{diff} = \mathbb{E}_{\epsilon, t, x_0}[||v_{\theta}(x_t, t) - v(\epsilon)||_2^2]$. 
With the same training pipeline as in Figure~\ref{fig:framewok}, the extension of ADT for flow-matching can be implemented by specifying the mechanism of the backward-flowing path constraint. 
Specifically, given the inference path $1=t_m>\ldots > t_0= 0 $, each inference step is conducted by:
\begin{equation}
    \begin{split}
        x_{t_{k-1}} = x_{t_k} - (t_k-t_{k-1}) v(sg(x_{t_k}), t_k).
    \end{split}
\end{equation}

% flow matching 的主要思想
% 迭代更新方式和优化目标
% 

% \textbf{Consistency Distillation Models}~\citep{song2023consistency,luo2023latent} speed the inference process of the diffusion model by distilling a neural network $f_{\theta}(x_t,t)$ for mapping the noisy data $x_t = \alpha_t x_0 + \sigma_t \epsilon$ to the clear data $x$, where $f_{\theta}(x_t,t)$ is set as the linear combination between the pretrained diffusion model $\epsilon_{\theta}(x_t,t)$ and $x_t$.
% With the original CM sampling strategy and the inference path $T>t_m>\ldots>t_0=0$, we can specify the sampling step with gradient flow contain as $x_{t_{k-1}} = \alpha_{t_k} f_{\theta}(sg(x_{t_{k}}), t_k) -sg(x_{t_{k}}) + x_{t_{k}} + \sigma_t\epsilon$.
% And the diffusion optimization loss is $\mathcal{L}_{diff} = \mathbb{E}_{x_0, k, \epsilon} [||f_{\theta}(x_{t_k}, t_k) - f_{\theta^{-}}(\hat{x}_{t_{k-1}}, t_{k-1})  ||_2^2]$, where $\hat{x}_{t_{k-1}}$ is the result of the one inference step from $x_{t_{k}}$, and $\theta^{-}$ is updated by the  exponential moving average of the parameter $\theta$.

\vspace{-3mm}
\section{Other Related Works and Comparisons }
\vspace{-1mm}
Technically, we stimulate the diffusion inference process during optimization and use adversarial supervision to close the generative and training distribution. 
In the literature, there exist some works that use similar technical parts.

\noindent\textbf{Diffusion Optimization with Adversarial Objective.} 
% Considering the powerful ability of adversarial training on distribution matching, 
Some recent works have tried to add the auxiliary adversarial loss between the final output and real data when optimizing the diffusion models~\citep{sauer2023adversarial}. However, due to the potential gradient back-propagation problem, all of them predict the final output with only one or a few model forwards~\citep{zhang2024unifl,ren2024byteedit}. Meanwhile, to ensure the fidelity, the model forward should start near the end of the inference path~\cite{ren2024byteedit}, or the base model should be distilled to enhance the prediction accuracy for the early inference step, like ADD~\citep{sauer2023adversarial}, UFOGen~\citep{xu2024ufogen}. 
They all leverage an approximation of the practical generative distribution and lead to sub-optimal optimization results. In contrast, ADT introduces the whole inference procedure into the training process for stimulating the practical generative distribution.
In particular, DMD2~\cite{yin2024improved} also stimulates the whole inference process during training. However, they focus on distilling diffusion models for accelerating sampling, where only 4 inference steps are considered during training. 
Here, we focus on improving the performance of the foundational diffusion models, which can also be further optimized or distilled for various downstream applications.
% (See the Appendix for more discussion)

\noindent\textbf{Diffusion Optimization with the Inference Procedure.}
It is crucial to back-propagate through the iterative diffusion inference process when optimizing the assessment of the final generative images. 
Several works avoid the direct model parameters updating by optimizing the sampler parameters or input noise~\citep{watson2022learning,wallace2023end}. Other works use reinforcement learning to fine-tune model parameters without flowing gradients through the inference path~\citep{black2023training,shen2023finetuning,lee2023aligning}. Recently, researchers have also turned to constraining the backward-flowing path
% to alleviate the risk of computational overhead and gradient explosion 
% when gradient back-propagating
~\citep{clark2023directly,prabhudesai2023aligning,wu2024deep}.
However, they all rely on robust reward models on images or labor-intensive human annotations to guide the optimization direction.
Unlike previous methods, ADT can operate on the common image dataset without additional annotations. The discriminator acts as a reward model to assess image similarity to the training data, dynamically updating to reduce the risk of over-exploitation seen in previous approaches~\citep{prabhudesai2023aligning,wu2024deep}.

% \begin{table}[t]
% \begin{tabular}{lcclll} \hline\hline
%                    & $a_{t_k}$  &  $b_{t_k}$ &  &  &  \\ \hline
% DDPM-eps           & $\frac{\alpha_{t_{k-1}}}{\alpha_{t_k}}$ & $ (\sigma_{t_{k-1}} - \sigma_{t_k}\frac{\alpha_{t_{k-1}}}{\alpha_{t_k}})$  &  &  &  \\ \hline
% Flow-matching      &     &  &  &  &  \\\hline
% Concisitency Model &    &  &  &  &  \\ \hline
% \end{tabular}
% \end{table}

% The generalization of ADM 
% LCM 
% Lora
% FM 

% Comparing with ADD UFOGen 
% Comparing with reward-based FT 
\begin{table*}[t]
\centering
\caption{The overall quantitative performance of the ADT framework.}\label{tab:overall_performance}
% \vspace{-2mm}
\resizebox{0.93\textwidth}{!}{%
\setlength\tabcolsep{4pt} 
\begin{tabular}{c|ccc|ccc|ccc|ccc|ccc}\hline\hline
     & \multicolumn{3}{c|}{SD15+DDIM50} & \multicolumn{3}{c|}{SD15+DPMS30} & \multicolumn{3}{c|}{SDXL+DDIM50} & \multicolumn{3}{c|}{SDXL+DPMS30} & \multicolumn{3}{c}{SD3} \\\cline{2-16}
     & FID$\downarrow$       & HPS$\uparrow$      & AES$\uparrow$      & FID$\downarrow$     & HPS$\uparrow$      & AES$\uparrow$      & FID$\downarrow$       & HPS$\uparrow$      & AES$\uparrow$      & FID$\downarrow$      & HPS$\uparrow$       & AES$\uparrow$       & FID$\downarrow$    & HPS$\uparrow$    & AES$\uparrow$    \\\hline
Base & 24.20     & 27.41    & 5.422    & 25.67    & 26.86    & 5.384    & 16.16     & 27.66    & 5.911    & 16.82    & 27.71    & 6.013    & 23.32  & 28.24  & 5.820 \\
FT   & 16.89     & 27.37    & 5.821    & 19.26    & 27.39    & 5.827    & 12.01     &  27.94    & 5.967   & 12.35  & 27.96   & 6.082    & 14.74 & 28.17  & 5.948 \\\hline
ADT  & \textbf{13.69}     & \textbf{27.59}    & \textbf{5.952}    & \textbf{14.65}    & \textbf{27.67}    & \textbf{5.910}    & \textbf{11.57}     &   \textbf{28.01}   & \textbf{6.110}     & \textbf{12.07}    & \textbf{28.07}   & \textbf{6.122}    & \textbf{13.49}  & \textbf{28.26}  & \textbf{5.963}\\
\hline\hline
\end{tabular}}
\vspace{-4mm}
\end{table*}
% \begin{figure*}[t]
%     \centering
%     % \subfloat[SD15+DDIM50]{\includegraphics[width=0.19\linewidth]{figs/sd15_ddim_score_rank.jpg}}
%     % \subfloat[SD15+DPM30]{\includegraphics[width=0.19\linewidth]{figs/sd15_dpm_score_rank.jpg}}
%     % \subfloat[SDXL+DDIM50]{\includegraphics[width=0.19\linewidth]{figs/sdxl_ddim_score_rank.jpg}}
%     % \subfloat[SDXL+DPM30]{\includegraphics[width=0.19\linewidth]{figs/sdxl_dpm_score_rank.jpg}}
%     % \subfloat[SD3]{\includegraphics[width=0.19\linewidth]{figs/sd3_score_rank.jpg}}
%     \includegraphics[width=0.19\linewidth]{figs/sd15_ddim_score_rank.jpg}
%     \includegraphics[width=0.19\linewidth]{figs/sd15_dpm_score_rank.jpg}
%     \includegraphics[width=0.19\linewidth]{figs/sdxl_ddim_score_rank.jpg}
%     \includegraphics[width=0.19\linewidth]{figs/sdxl_dpm_score_rank.jpg}
%     \includegraphics[width=0.19\linewidth]{figs/sd3_score_rank.jpg}\vspace{-2mm}
%     \subfloat[SD15+DDIM50]{\includegraphics[width=0.19\linewidth]{figs/sd15_ddim_human_rank.jpg}}
%     \subfloat[SD15+DPMS30]{\includegraphics[width=0.19\linewidth]{figs/sd15_dpm_human_rank.jpg}}
%     \subfloat[SDXL+DDIM50]{\includegraphics[width=0.19\linewidth]{figs/sdxl_ddim_human_rank.jpg}}
%     \subfloat[SDXL+DPMS30]{\includegraphics[width=0.19\linewidth]{figs/sdxl_dpm_human_rank.jpg}}
%     \subfloat[SD3]{\includegraphics[width=0.19\linewidth]{figs/sd3_human_rank.jpg}}
%     \vspace{-2mm}
%     \caption{The win rate of different training strategies ranked by quality assessment models (Top) and human experts (Bottom).}
%     \vspace{-4mm}
%     \label{fig:score_rank}
% \end{figure*}

\begin{figure*}
    \centering
    \includegraphics[width=0.93\linewidth]{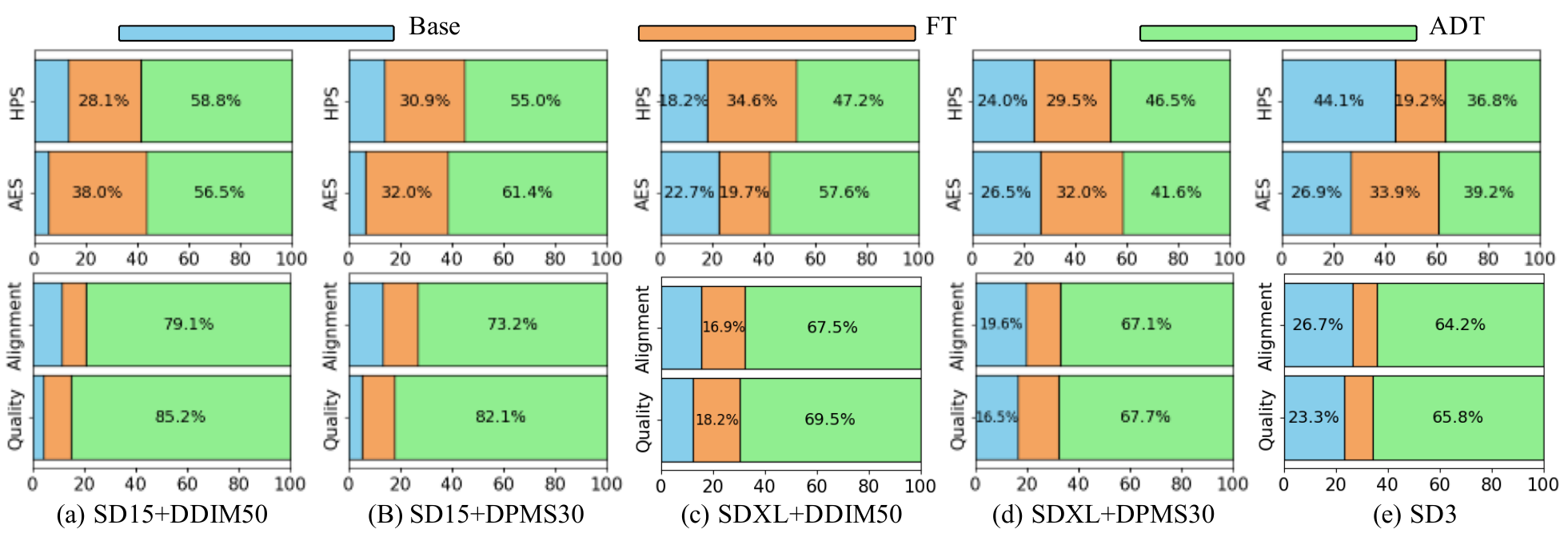}
    \vspace{-4mm}
    \caption{The win rate of different training strategies ranked by quality assessment models (Top) and human experts (Bottom).}
    \vspace{-4mm}
    \label{fig:score_rank}
\end{figure*}

\vspace{-2mm}
\section{Experiments}
\vspace{-1mm}

\begin{figure*}[t]
    \centering
    \includegraphics[width=0.95\linewidth]{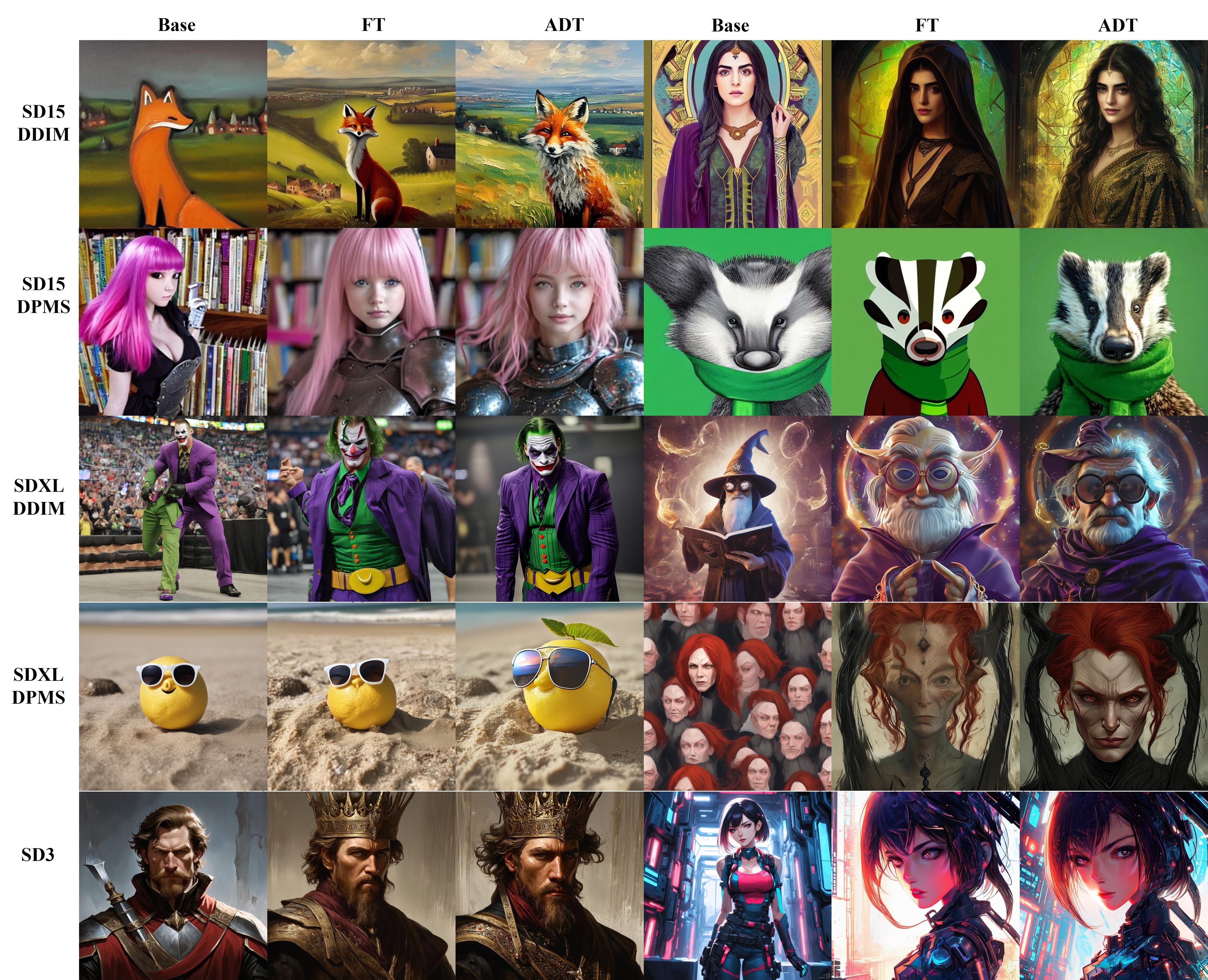}
    \vspace{-2mm}
    \caption{The generated samples under different training strategies, where captions sampled from HPSV2 Benchmark prompts.}
    \vspace{-4mm}
    \label{fig:cases}
\end{figure*}
%SD 1.5 SDXL  SD3
\noindent\textbf{Benchmark Models.}
We introduce two diffusion models and one flow-matching model as base models, which have different network backbones and parameter scales: 
1) Stable Diffusion v 1.5 (\textbf{SD15})\cite{rombach2022high} is the typical text-to-image diffusion model, which applies the U-net~\cite{ronneberger2015u} with  0.86B parameters as the diffusion backbone; 
2) Stable Diffusion XL  base model (\textbf{SDXL})~\cite{podell2023sdxl} is the extended version of SD with overall 3.5B parameters. In particular, SDXL supports generating high-resolution images at 1024 resolution. 
3) Stable Diffusion 3 medium (\textbf{SD3})~\cite{esser2024scaling} is a large-scale flow matching model with transformers-based backbone, i.e., DiT~\cite{peebles2023scalable} with overall 2B parameters.
SD15 can generate images of 512 resolution, while  SDXL and SD3 support images at 1024 resolution. 
% combining CNNs and Cross-attention Layers~\cite{radford2021learning}
%more cross-attention layers and
% for modeling both text and image modalities and

%Sampler CFG 
Meanwhile, for SD15 and SDXL, two samplers are discussed, i.e., DDIM~\cite{song2020denoising} with 50 diffusion steps, and DPMSlover++~\cite{lu2022dpm2} (DPMS) with 30 diffusion steps, which are both the most widely used in practice. 
As for SD3, the original flow-matching sampler is used with 28 steps. 
The CFG strategy~\cite{ho2022classifier} is also used for all three models with different guidance degrees, i.e., 7.0 for SD15 and SD3, and 5.0 for SDXL.

\noindent\textbf{Dataset.} 
To assess ADT's ability to align generated and training data distributions, we increase optimization challenges by using training samples not included in the pre-training dataset.
% To evaluate our ADT framework on closing the generated and training data distribution, we hope to increase the optimization challenge by constructing the training samples outside the pre-training dataset. 
Along this line, we follow JourneyDB dataset~\cite{sun2023journeydb} and collect 170K high-resolution and high-quality text-image pairs from Midjourney v6~\cite{midjourbey}, a SOTA text-to-image generation online platform and released after the publication date of all three benchmark models. In particular, it commonly outputs high-quality images. Meanwhile, 10k text-image pairs are randomly selected for the test dataset if needed, while others are for training. Numerical statistics can be found in the Appendix.

\noindent\textbf{Metric.} We evaluate ADT with three types of metrics: 
1) Firstly, following ~\cite{goodfellow2020generative,rombach2022high}, we use the Frechet Inception Distance Score (\textbf{FID}) between the generative and the test datasets as the direct metric of the divergence of generative and training distributions. 
% Specifically, 10k images are randomly generated by the tuned models with the same prompts as the test dataset.
2) Secondly, generated images ideally match the quality of the training data when their distributions align. 
With high-quality images as training data, the trained models should also output high-quality images.
We used two image quality assessment models for evaluation: HPSv2 Score (\textbf{HPS})~\cite{wu2023human2}, and Aesthetics Score (\textbf{AES})~\cite{Christoph2022aes}, which are trained to align the human preference with different network architectures. The HPSv2 benchmark prompts are used for the test, including 3200 captions with diverse styles.
3) Thirdly, \textbf{Human-level evaluation} is also involved in assessing whether quality truly improves after training on high-quality datasets

\noindent\textbf{Training Settings.}
We use AdamW~\cite{kingma2014adam} as the optimizer for both the discriminator and generator. The whole training batch size is set as 8 with 4 NVIDIA A100 GPUs. We train at fixed square resolutions, i.e., 512 for SD and 1024 for SDXL and SD3. The learning rate is set as the same for the discriminator and generator, 5e-6 for SD and 1e-6 for SDXL and SD3. The linear warmup is used with 1000 steps for the discriminator and 500 steps for the generator.  The generator updating frequency is 1/5. The EMA rate $\tau$ is set as 0.99. The weight parameter $\lambda$ is set as 0.5.  The hyper-parameter $K$ is set as 3. 
% More demos and part of code can be found at \url{http://www.XXXX.com}
Our code will be publicly accessible after acceptance.

% \usepackage{multirow}
% \usepackage{multirow}

% \begin{figure*}
%     \centering
%     \subfloat[SD15+DDIM50]{\includegraphics[width=0.19\linewidth]{figs/sd15_ddim_human_rank.jpg}}
%     \subfloat[SD15+DPM30]{\includegraphics[width=0.19\linewidth]{figs/sd15_dpm_human_rank.jpg}}
%     \subfloat[SDXL+DDIM50]{\includegraphics[width=0.19\linewidth]{figs/sdxl_ddim_human_rank.jpg}}
%     \subfloat[SDXL+DPM30]{\includegraphics[width=0.19\linewidth]{figs/sdxl_dpm_human_rank.jpg}}
%     \subfloat[SD3]{\includegraphics[width=0.19\linewidth]{figs/sd3_human_rank.jpg}}
%     \vspace{-3mm}
%     \caption{The win rate of different training strategies ranking by human.}
%     \vspace{-4mm}
%     \label{fig:human_rank}
% \end{figure*}

% \subsection{Distribution Matching }

% \subsection{Qualtity Assessment}

% \subsection{Human-level Evaluation}

\vspace{-1mm}
\subsection{Overall Performance}
\vspace{-2mm}
Table~\ref{tab:overall_performance} shows the overall quantitative performance of our ADT framework, where the original fine-tuning strategy (FT) is used as the baseline. Both ADT and FT are trained with the same number of training iterations, i.e. 20k. The original base model is also involved for clear comparison.  Our ADT framework is trained with the same sampler setting used for the inference.
For more fairness comparisons, we also illustrate the win rate for different training frameworks in terms of HPS and AES scores in Figure~\ref{fig:score_rank}.

We can find that ADT achieves the best performance among all settings, which demonstrates that ADT can both close the distribution distance between training and generated data, and align the image quality with the training data. 
Interestingly, ADT achieves large improvement in the small model, i.e., SD15, than in the models with large parameter scales, i.e., SDXL and SD3, even reaching similar FID scores between SD15 and SD3.
This may demonstrate that ADT can inspire the model to realize greater parameter utilization efficiency.
In addition, SDXL-based models achieve better FID scores, as their base model's generated data more closely aligns with our dataset. In contrast, the style of generated data for SD3 is largely different from our dataset, where the outputs of SD3 are more like real photos while our dataset from Midjourney is like painting.
As a result, the FT strategy may not narrow this distribution gap well and perform worse in terms of HPS scores, while ADT can make up for this setback and achieve better performance.
In particular, we have checked that ADT captures 51\% of the pair-wise win rate compared with the base model.

Figure~\ref{fig:cases} displays qualitative comparisons among different models with different training strategies. ADT produces more appealing
imagery, with fine-grained details, vivid arrays of colors, and good composition.
To be specific, images from FT and ADT models both present similar color and lighting styles to our dataset, while ADM models can also create more image details on animal hair, facial wrinkles, or object texture, and enrich the image background with reasonable color variations and object filling.
Therefore, we further ask 5 experts to rank the generated images from models with different training strategies with 100 prompts randomly selected from HPS test prompts from two aspects: the image quality and text alignment. 
The results are summarized at the below line in Figure~\ref{fig:score_rank}.
ADT achieved an overwhelming victory on all three base models.
More qualitative can be found in the Appendix.

\vspace{-2mm}
\subsection{Generalization for Samplers}
\vspace{-1mm}
% In the experiments above, we demonstrated the robustness of our ADT framework across various models, scales, and samplers, using a consistent sampler for evaluation. 
Here, we further explore ADT's generalization with different sampler settings.
Specifically, we take the SD15 as an example and compare the performance of the trained model with different sampling settings or different samplers.

% \begin{figure}
%     \centering
%     \includegraphics[width=0.31\linewidth]{figs/cfg_fid.jpg}
%     \includegraphics[width=0.31\linewidth]{figs/cfg_hps.jpg}
%     \includegraphics[width=0.31\linewidth]{figs/cfg_aes.jpg}
%     \includegraphics[width=0.31\linewidth]{figs/ns_fid.jpg}
%     \includegraphics[width=0.31\linewidth]{figs/ns_hps.jpg}
%     \includegraphics[width=0.31\linewidth]{figs/ns_aes.jpg}
%     \vspace{-2mm}
%     \caption{The performance of ADT with different CFG scales (Top) or numbers of inference steps (Bottom).}
%     \vspace{-2mm}
%     \label{fig:sampler_cfg}
% \end{figure}

\begin{table}[t]
\centering
\caption{The performance of ADT when using an inference sampler different from the one used during training.}
% \vspace{-2mm}
\label{tab:sampler_sampler}
\resizebox{0.95\linewidth}{!}{%
\setlength\tabcolsep{3pt}  
\begin{tabular}{c|ccc|ccc} \hline\hline
         & \multicolumn{3}{c|}{DDIM Sampler}        & \multicolumn{3}{c}{DPMS Sampler} \\\cline{2-7}
         & FID-5k         & HPS   & AES            & FID-5k & HPS            & AES   \\\hline
FT       & 23.08          & 27.37 & 5.827          & 25.38   & 27.39          & 5.827 \\ \hline
ADT-DDIM & \textbf{19.84} & 27.60 & \textbf{5.952} & 20.73  & 27.67          & 5.925 \\
ADT-DPMS  & 20.03          & 27.59 & 5.910          & 20.83  & \textbf{27.68} & 5.910 \\
\hline\hline
\end{tabular}}
\vspace{-3mm}
\end{table}

\begin{figure}
    \centering
    \includegraphics[width=0.95\linewidth]{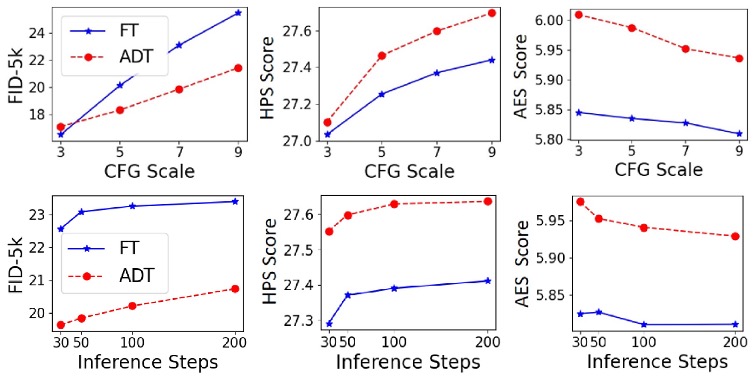}
    \vspace{-3mm}
    \caption{The performance of ADT with different CFG scales (Top) or numbers of inference steps (Bottom).}
    \vspace{-6mm}
    \label{fig:sampler_cfg}
\end{figure}

% \begin{figure}
%     \centering
%     \includegraphics[width=0.31\linewidth]{figs/ns_fid.jpg}
%     \includegraphics[width=0.31\linewidth]{figs/ns_hps.jpg}
%     \includegraphics[width=0.31\linewidth]{figs/ns_aes.jpg}
%     \caption{The performance of ADT with different CFG scales, using the DDIM sampler with the CFG scale of 7.0.}
%     \label{fig:sampler_ns}
% \end{figure}

\noindent\textbf{Different Sampling Settings.}
Using the DDIM sampler, we first fixed the inference steps as 50 and,  vary the CFG scales among [3.0,5.0,7.0,9.0]. The assessment of generated images is shown in the top line of Figure~\ref{fig:sampler_cfg}. We can find our ADT framework can mostly achieve better performance compared with the original fine-tuning strategy with a lower FID score, and higher quality scores, i.e., HPS and  AES scores. In addition, we can find that ADT can achieve a slower FID increase and faster HPS score increase when the CFG scale grows.
Second, by fixing the CFG scale as 7.0, we vary the inference steps among [30, 50, 100, 200] and show the results in the bottom line of Figure~\ref{fig:sampler_cfg}. Our ADT framework has achieved a large performance improvement compared with FT in all settings with similar trends of change, where the performance will converge with 100 inference steps or more steps. All observations demonstrate the generalization of our ADT on varying sampling settings.

\vspace{-1mm}
\noindent\textbf{Different Sampling Methods.}
We further verify the generalization by using inconsistent inference samplers from those used during training. Table~\ref{tab:sampler_sampler} presents the results. We observe that consistent inference samplers enable models to achieve the best scores across several metrics. For instance, ADT-DDIM with the DDIM sampler attains the best FID and AES scores, while ADT-DPMS with the DPMS sampler achieves the highest HPS score. Additionally, inconsistent samplers can yield similar results. The image quality may be less sensitive to the samplers used during training than the inference samplers.

\vspace{-2mm}
\subsection{Ablation Analysis}\label{sec:ablation}
\vspace{-1mm}
Here, we conduct the ablation analysis with three variants of our ADT framework: 
% \begin{itemize}
1) \textbf{+ StyleGAN}  replace our discriminator with the discriminator used in StyleGAN-T~\cite{sauer2023stylegan}, which also applies the pre-trained DINOv2~\cite{oquab2023dinov2} as the backbone without the siamese-network-based discriminator heads.
2) \textbf{+ DRTune} conduct the gradient backpropagation following  DRtune~\cite{wu2024deep} when the generator updating.
3) \textbf{w/o $\mathcal{L}_{diff}$} removes the original diffusion loss during training.
% \end{itemize}

Take the SD15 as the example, we summarize the results in Table~\ref{tab:ablation}. We can find that our ADT has achieved the best performance with a large score margin compared with all variants. 
To be specific, removing the original diffusion loss significantly impairs the performance of distribution matching a lot, where the discriminator may be hacked.
In addition, our siamese-network-based discriminator demonstrates superior performance compared to StyleGAN-based variants, while our constraints on the backward-flowing path are more effective than DRTune.
% Compared with ADT with StyleGAN, we can conclusion that our siamese-network-based discriminator provides more contributions to the generated images. Similarly, compared with ADT with DRTune, our gradient flow constrain algorithm can achieve a better effect.
% \textit{Whether the Siamese-Network Based Discriminator matters? } 
% \textit{Whether our gradient flow constraint is better than DR-Tune?}
% \textit{Whether $L_{diff}$ is necessary?}
% To be specific, three variants of our ADT framework have been involved: 1) \textbf{w StyleGAN} replaces our discriminator with the discriminator used in StyleGAN2~\cite{}, which also apply the pre-trained DINOV2 as the backbone without the siamese-network based discriminator heads.
% 2) \textbf{w DRTune} conduct the gradient backpropagation following  DRtune~\cite{} when the generator updating.
% 3) 

% 4) What  is the best choice of the number of training timesteps $K$

\begin{table}[t]
\centering
\caption{The ablation analysis.}\label{tab:ablation}
% \vspace{-2mm}
\resizebox{0.95\linewidth}{!}{%
\setlength\tabcolsep{3pt} 
\begin{tabular}{c|c|cc|cc} \hline \hline
                         &        & \multicolumn{2}{c|}{HPS Score} & \multicolumn{2}{c}{AES Score} \\ \cline{2-6}
                         & FID-5k & Score        & Win Rate       & Score        & Win Rate       \\ \hline
ADT                      & \textbf{19.84}  & \textbf{27.59}        & N/A          & \textbf{5.952}        &  N/A           \\ \hline
+ StyleGAN               & 22.35  & 27.48        & 0.5587         & 5.838        & 0.6381         \\
+ DRTune                 & 21.92  & 27.40        & 0.6014         & 5.849        & 0.6594     \\
w/o $\mathcal{L}_{diff}$ & 24.03  & 27.18        & 0.7778         & 5.873        & 0.5993         \\
\hline \hline
\end{tabular}}
\vspace{-5mm}
\end{table}

% FID  HPS score AES score pick Score 
% generalization/ robustness 
% ablation 

%看一下CVPR 2024论文的实验室设置

\vspace{-2mm}
\section{Conclusions}
\vspace{-2mm}

We propose Adversarial Diffusion Tuning (ADT), a fine-tuning framework that addresses training-inference divergence in diffusion models.
The core idea is incorporating the inference process into optimization and using adversarial objectives to align generative with training data. For robust adversarial supervision, ADT employs a siamese-network discriminator with a pre-trained backbone, utilizes an image-to-image sampling strategy, and preserves the original diffusion loss. We develop a backward-flowing path constraint mechanism to prevent memory overload and gradient explosion. Experiments across various diffusion models demonstrate ADT's effectiveness.

% In this paper, we propose an effective fine-tuning framework for diffusion models, called Adversarial Diffusion Tuning (ADT), to alleviate the training-inference divergence in the original diffusion training process. The main idea is to integrate the whole inference process into the optimization procedure and use adversarial objectives to align the final outputs with real data. Specifically, to achieve robust adversarial training,  ADT features a siamese-network discriminator with a pre-trained backbone to enhance discriminative capabilities, incorporates an image-to-image sampling strategy to smooth discriminative difficulties, and preserves the original diffusion loss to prevent discriminator hacking. In addition, to avoid memory overload and gradient explosion during gradient back-propagation, an efficient gradient flow constrain mechanism is developed. Extensive experiments on diverse diffusion models with different model formulations, network architectures, and parameter scales, demonstrate the effectiveness of ADT on both distribution alignment and image quality improvement.

%%%介绍时间复杂度上缺陷

\section{Impact Statemente}
This paper presents work whose goal is to advance the field of Computer Vision and Machine Learning. There are many potential societal consequences of our work, none of which we feel must be specifically highlighted.

\bibliography{cvpr2025}

\begin{thebibliography}{54}
\providecommand{\natexlab}[1]{#1}
\providecommand{\url}[1]{\texttt{#1}}
\expandafter\ifx\csname urlstyle\endcsname\relax
  \providecommand{\doi}[1]{doi: #1}\else
  \providecommand{\doi}{doi: \begingroup \urlstyle{rm}\Url}\fi

\bibitem[flu()]{flux}
Flux.1-dev.
\newblock \url{https://huggingface.co/black-forest-labs/FLUX.1-dev}.
\newblock Accessed: 2024 - 11- 14.

\bibitem[mid()]{midjourbey}
Midjourney.
\newblock \url{https://www.midjourney.com/home}.
\newblock Accessed: 2024 - 11- 14.

\bibitem[Bai et~al.(2021)Bai, Luo, Zhao, Wen, and Wang]{bai2021recent}
Bai, T., Luo, J., Zhao, J., Wen, B., and Wang, Q.
\newblock Recent advances in adversarial training for adversarial robustness.
\newblock \emph{arXiv preprint arXiv:2102.01356}, 2021.

\bibitem[Bao et~al.(2022)Bao, Li, Zhu, and Zhang]{bao2022analytic}
Bao, F., Li, C., Zhu, J., and Zhang, B.
\newblock Analytic-dpm: an analytic estimate of the optimal reverse variance in diffusion probabilistic models.
\newblock \emph{arXiv preprint arXiv:2201.06503}, 2022.

\bibitem[Black et~al.(2023)Black, Janner, Du, Kostrikov, and Levine]{black2023training}
Black, K., Janner, M., Du, Y., Kostrikov, I., and Levine, S.
\newblock Training diffusion models with reinforcement learning.
\newblock \emph{arXiv preprint arXiv:2305.13301}, 2023.

\bibitem[Chen(2023)]{chen2023importance}
Chen, T.
\newblock On the importance of noise scheduling for diffusion models.
\newblock \emph{arXiv preprint arXiv:2301.10972}, 2023.

\bibitem[Chen et~al.(2020)Chen, Kornblith, Norouzi, and Hinton]{chen2020simple}
Chen, T., Kornblith, S., Norouzi, M., and Hinton, G.
\newblock A simple framework for contrastive learning of visual representations.
\newblock In \emph{International conference on machine learning}, pp.\  1597--1607. PMLR, 2020.

\bibitem[Clark et~al.(2023)Clark, Vicol, Swersky, and Fleet]{clark2023directly}
Clark, K., Vicol, P., Swersky, K., and Fleet, D.~J.
\newblock Directly fine-tuning diffusion models on differentiable rewards.
\newblock \emph{arXiv preprint arXiv:2309.17400}, 2023.

\bibitem[Dosovitskiy et~al.(2020)Dosovitskiy, Beyer, Kolesnikov, Weissenborn, Zhai, Unterthiner, Dehghani, Minderer, Heigold, Gelly, et~al.]{dosovitskiy2020image}
Dosovitskiy, A., Beyer, L., Kolesnikov, A., Weissenborn, D., Zhai, X., Unterthiner, T., Dehghani, M., Minderer, M., Heigold, G., Gelly, S., et~al.
\newblock An image is worth 16x16 words: Transformers for image recognition at scale.
\newblock \emph{arXiv preprint arXiv:2010.11929}, 2020.

\bibitem[Esser et~al.(2024)Esser, Kulal, Blattmann, Entezari, M{\"u}ller, Saini, Levi, Lorenz, Sauer, Boesel, et~al.]{esser2024scaling}
Esser, P., Kulal, S., Blattmann, A., Entezari, R., M{\"u}ller, J., Saini, H., Levi, Y., Lorenz, D., Sauer, A., Boesel, F., et~al.
\newblock Scaling rectified flow transformers for high-resolution image synthesis.
\newblock In \emph{Forty-first International Conference on Machine Learning}, 2024.

\bibitem[Goodfellow et~al.(2020)Goodfellow, Pouget-Abadie, Mirza, Xu, Warde-Farley, Ozair, Courville, and Bengio]{goodfellow2020generative}
Goodfellow, I., Pouget-Abadie, J., Mirza, M., Xu, B., Warde-Farley, D., Ozair, S., Courville, A., and Bengio, Y.
\newblock Generative adversarial networks.
\newblock \emph{Communications of the ACM}, 63\penalty0 (11):\penalty0 139--144, 2020.

\bibitem[Grill et~al.(2020)Grill, Strub, Altch{\'e}, Tallec, Richemond, Buchatskaya, Doersch, Avila~Pires, Guo, Gheshlaghi~Azar, et~al.]{grill2020bootstrap}
Grill, J.-B., Strub, F., Altch{\'e}, F., Tallec, C., Richemond, P., Buchatskaya, E., Doersch, C., Avila~Pires, B., Guo, Z., Gheshlaghi~Azar, M., et~al.
\newblock Bootstrap your own latent-a new approach to self-supervised learning.
\newblock \emph{Advances in neural information processing systems}, 33:\penalty0 21271--21284, 2020.

\bibitem[He et~al.(2020)He, Fan, Wu, Xie, and Girshick]{he2020momentum}
He, K., Fan, H., Wu, Y., Xie, S., and Girshick, R.
\newblock Momentum contrast for unsupervised visual representation learning.
\newblock In \emph{Proceedings of the IEEE/CVF conference on computer vision and pattern recognition}, pp.\  9729--9738, 2020.

\bibitem[Ho \& Salimans(2022)Ho and Salimans]{ho2022classifier}
Ho, J. and Salimans, T.
\newblock Classifier-free diffusion guidance.
\newblock \emph{arXiv preprint arXiv:2207.12598}, 2022.

\bibitem[Ho et~al.(2020)Ho, Jain, and Abbeel]{ho2020denoising}
Ho, J., Jain, A., and Abbeel, P.
\newblock Denoising diffusion probabilistic models.
\newblock \emph{Advances in neural information processing systems}, 33:\penalty0 6840--6851, 2020.

\bibitem[Karras et~al.(2022)Karras, Aittala, Aila, and Laine]{karras2022elucidating}
Karras, T., Aittala, M., Aila, T., and Laine, S.
\newblock Elucidating the design space of diffusion-based generative models.
\newblock \emph{Advances in neural information processing systems}, 35:\penalty0 26565--26577, 2022.

\bibitem[Kingma(2014)]{kingma2014adam}
Kingma, D.~P.
\newblock Adam: A method for stochastic optimization.
\newblock \emph{arXiv preprint arXiv:1412.6980}, 2014.

\bibitem[Kingma \& Welling(2013)Kingma and Welling]{kingma2013auto}
Kingma, D.~P. and Welling, M.
\newblock Auto-encoding variational bayes.
\newblock \emph{arXiv preprint arXiv:1312.6114}, 2013.

\bibitem[Kirstain et~al.(2023)Kirstain, Polyak, Singer, Matiana, Penna, and Levy]{kirstain2023pick}
Kirstain, Y., Polyak, A., Singer, U., Matiana, S., Penna, J., and Levy, O.
\newblock Pick-a-pic: An open dataset of user preferences for text-to-image generation.
\newblock \emph{Advances in Neural Information Processing Systems}, 36:\penalty0 36652--36663, 2023.

\bibitem[Lee et~al.(2023)Lee, Liu, Ryu, Watkins, Du, Boutilier, Abbeel, Ghavamzadeh, and Gu]{lee2023aligning}
Lee, K., Liu, H., Ryu, M., Watkins, O., Du, Y., Boutilier, C., Abbeel, P., Ghavamzadeh, M., and Gu, S.~S.
\newblock Aligning text-to-image models using human feedback.
\newblock \emph{arXiv preprint arXiv:2302.12192}, 2023.

\bibitem[Li et~al.(2023)Li, Qu, Yao, Sun, and Moens]{lialleviating}
Li, M., Qu, T., Yao, R., Sun, W., and Moens, M.-F.
\newblock Alleviating exposure bias in diffusion models through sampling with shifted time steps.
\newblock In \emph{The Twelfth International Conference on Learning Representations}, 2023.

\bibitem[Lim \& Ye(2017)Lim and Ye]{lim2017geometric}
Lim, J.~H. and Ye, J.~C.
\newblock Geometric gan.
\newblock \emph{arXiv preprint arXiv:1705.02894}, 2017.

\bibitem[Liu et~al.(2022)Liu, Gong, and Liu]{liu2022flow}
Liu, X., Gong, C., and Liu, Q.
\newblock Flow straight and fast: Learning to generate and transfer data with rectified flow.
\newblock \emph{arXiv preprint arXiv:2209.03003}, 2022.

\bibitem[Lu et~al.(2022{\natexlab{a}})Lu, Zhou, Bao, Chen, Li, and Zhu]{lu2022dpm}
Lu, C., Zhou, Y., Bao, F., Chen, J., Li, C., and Zhu, J.
\newblock Dpm-solver: A fast ode solver for diffusion probabilistic model sampling in around 10 steps.
\newblock \emph{Advances in Neural Information Processing Systems}, 35:\penalty0 5775--5787, 2022{\natexlab{a}}.

\bibitem[Lu et~al.(2022{\natexlab{b}})Lu, Zhou, Bao, Chen, Li, and Zhu]{lu2022dpm2}
Lu, C., Zhou, Y., Bao, F., Chen, J., Li, C., and Zhu, J.
\newblock Dpm-solver++: Fast solver for guided sampling of diffusion probabilistic models.
\newblock \emph{arXiv preprint arXiv:2211.01095}, 2022{\natexlab{b}}.

\bibitem[Murray et~al.(2012)Murray, Marchesotti, and Perronnin]{murray2012ava}
Murray, N., Marchesotti, L., and Perronnin, F.
\newblock Ava: A large-scale database for aesthetic visual analysis.
\newblock In \emph{2012 IEEE conference on computer vision and pattern recognition}, pp.\  2408--2415. IEEE, 2012.

\bibitem[Nichol \& Dhariwal(2021)Nichol and Dhariwal]{nichol2021improved}
Nichol, A.~Q. and Dhariwal, P.
\newblock Improved denoising diffusion probabilistic models.
\newblock In \emph{International conference on machine learning}, pp.\  8162--8171. PMLR, 2021.

\bibitem[Ning et~al.(2023{\natexlab{a}})Ning, Li, Su, Salah, and Ertugrul]{ningelucidating}
Ning, M., Li, M., Su, J., Salah, A.~A., and Ertugrul, I.~O.
\newblock Elucidating the exposure bias in diffusion models.
\newblock In \emph{The Twelfth International Conference on Learning Representations}, 2023{\natexlab{a}}.

\bibitem[Ning et~al.(2023{\natexlab{b}})Ning, Sangineto, Porrello, Calderara, and Cucchiara]{ning2023input}
Ning, M., Sangineto, E., Porrello, A., Calderara, S., and Cucchiara, R.
\newblock Input perturbation reduces exposure bias in diffusion models.
\newblock In \emph{International Conference on Machine Learning}, pp.\  26245--26265. PMLR, 2023{\natexlab{b}}.

\bibitem[Oquab et~al.(2023)Oquab, Darcet, Moutakanni, Vo, Szafraniec, Khalidov, Fernandez, Haziza, Massa, El-Nouby, et~al.]{oquab2023dinov2}
Oquab, M., Darcet, T., Moutakanni, T., Vo, H., Szafraniec, M., Khalidov, V., Fernandez, P., Haziza, D., Massa, F., El-Nouby, A., et~al.
\newblock Dinov2: Learning robust visual features without supervision.
\newblock \emph{arXiv preprint arXiv:2304.07193}, 2023.

\bibitem[Peebles \& Xie(2023)Peebles and Xie]{peebles2023scalable}
Peebles, W. and Xie, S.
\newblock Scalable diffusion models with transformers.
\newblock In \emph{Proceedings of the IEEE/CVF International Conference on Computer Vision}, pp.\  4195--4205, 2023.

\bibitem[Podell et~al.(2023)Podell, English, Lacey, Blattmann, Dockhorn, M{\"u}ller, Penna, and Rombach]{podell2023sdxl}
Podell, D., English, Z., Lacey, K., Blattmann, A., Dockhorn, T., M{\"u}ller, J., Penna, J., and Rombach, R.
\newblock Sdxl: Improving latent diffusion models for high-resolution image synthesis.
\newblock \emph{arXiv preprint arXiv:2307.01952}, 2023.

\bibitem[Prabhudesai et~al.(2023)Prabhudesai, Goyal, Pathak, and Fragkiadaki]{prabhudesai2023aligning}
Prabhudesai, M., Goyal, A., Pathak, D., and Fragkiadaki, K.
\newblock Aligning text-to-image diffusion models with reward backpropagation.
\newblock \emph{arXiv preprint arXiv:2310.03739}, 2023.

\bibitem[Ren et~al.(2024)Ren, Wu, Lu, Kuang, Xia, Wang, Wang, Zhu, Xie, Wang, et~al.]{ren2024byteedit}
Ren, Y., Wu, J., Lu, Y., Kuang, H., Xia, X., Wang, X., Wang, Q., Zhu, Y., Xie, P., Wang, S., et~al.
\newblock Byteedit: Boost, comply and accelerate generative image editing.
\newblock \emph{arXiv preprint arXiv:2404.04860}, 2024.

\bibitem[Rombach et~al.(2022)Rombach, Blattmann, Lorenz, Esser, and Ommer]{rombach2022high}
Rombach, R., Blattmann, A., Lorenz, D., Esser, P., and Ommer, B.
\newblock High-resolution image synthesis with latent diffusion models.
\newblock In \emph{Proceedings of the IEEE/CVF conference on computer vision and pattern recognition}, pp.\  10684--10695, 2022.

\bibitem[Ronneberger et~al.(2015)Ronneberger, Fischer, and Brox]{ronneberger2015u}
Ronneberger, O., Fischer, P., and Brox, T.
\newblock U-net: Convolutional networks for biomedical image segmentation.
\newblock In \emph{Medical image computing and computer-assisted intervention--MICCAI 2015: 18th international conference, Munich, Germany, October 5-9, 2015, proceedings, part III 18}, pp.\  234--241. Springer, 2015.

\bibitem[Sauer et~al.(2023{\natexlab{a}})Sauer, Karras, Laine, Geiger, and Aila]{sauer2023stylegan}
Sauer, A., Karras, T., Laine, S., Geiger, A., and Aila, T.
\newblock Stylegan-t: Unlocking the power of gans for fast large-scale text-to-image synthesis.
\newblock In \emph{International conference on machine learning}, pp.\  30105--30118. PMLR, 2023{\natexlab{a}}.

\bibitem[Sauer et~al.(2023{\natexlab{b}})Sauer, Lorenz, Blattmann, and Rombach]{sauer2023adversarial}
Sauer, A., Lorenz, D., Blattmann, A., and Rombach, R.
\newblock Adversarial diffusion distillation.
\newblock \emph{arXiv preprint arXiv:2311.17042}, 2023{\natexlab{b}}.

\bibitem[Schuhmann(2022)]{Christoph2022aes}
Schuhmann, C.
\newblock Laion-aesthetics.
\newblock \url{https://laion.ai/blog/laion-aesthetics/}, 2022.
\newblock Accessed: 2023 - 11- 10.

\bibitem[Shen et~al.(2023)Shen, Du, Pang, Lin, Wong, and Kankanhalli]{shen2023finetuning}
Shen, X., Du, C., Pang, T., Lin, M., Wong, Y., and Kankanhalli, M.
\newblock Finetuning text-to-image diffusion models for fairness.
\newblock \emph{arXiv preprint arXiv:2311.07604}, 2023.

\bibitem[Sohl-Dickstein et~al.(2015)Sohl-Dickstein, Weiss, Maheswaranathan, and Ganguli]{sohl2015deep}
Sohl-Dickstein, J., Weiss, E., Maheswaranathan, N., and Ganguli, S.
\newblock Deep unsupervised learning using nonequilibrium thermodynamics.
\newblock In \emph{International conference on machine learning}, pp.\  2256--2265. PMLR, 2015.

\bibitem[Song et~al.(2020{\natexlab{a}})Song, Meng, and Ermon]{song2020denoising}
Song, J., Meng, C., and Ermon, S.
\newblock Denoising diffusion implicit models.
\newblock \emph{arXiv preprint arXiv:2010.02502}, 2020{\natexlab{a}}.

\bibitem[Song et~al.(2020{\natexlab{b}})Song, Sohl-Dickstein, Kingma, Kumar, Ermon, and Poole]{song2020score}
Song, Y., Sohl-Dickstein, J., Kingma, D.~P., Kumar, A., Ermon, S., and Poole, B.
\newblock Score-based generative modeling through stochastic differential equations.
\newblock \emph{arXiv preprint arXiv:2011.13456}, 2020{\natexlab{b}}.

\bibitem[Sun et~al.(2023)Sun, Pan, Ge, Li, Duan, Wu, Zhang, Zhou, Qin, Wang, et~al.]{sun2023journeydb}
Sun, K., Pan, J., Ge, Y., Li, H., Duan, H., Wu, X., Zhang, R., Zhou, A., Qin, Z., Wang, Y., et~al.
\newblock Journeydb: A benchmark for generative image understanding.
\newblock \emph{Advances in Neural Information Processing Systems}, 36, 2023.

\bibitem[Wallace et~al.(2023)Wallace, Gokul, Ermon, and Naik]{wallace2023end}
Wallace, B., Gokul, A., Ermon, S., and Naik, N.
\newblock End-to-end diffusion latent optimization improves classifier guidance.
\newblock In \emph{Proceedings of the IEEE/CVF International Conference on Computer Vision}, pp.\  7280--7290, 2023.

\bibitem[Watson et~al.(2022)Watson, Chan, Ho, and Norouzi]{watson2022learning}
Watson, D., Chan, W., Ho, J., and Norouzi, M.
\newblock Learning fast samplers for diffusion models by differentiating through sample quality.
\newblock In \emph{International Conference on Learning Representations}, 2022.

\bibitem[Wu et~al.(2023{\natexlab{a}})Wu, Hao, Sun, Chen, Zhu, Zhao, and Li]{wu2023human2}
Wu, X., Hao, Y., Sun, K., Chen, Y., Zhu, F., Zhao, R., and Li, H.
\newblock Human preference score v2: A solid benchmark for evaluating human preferences of text-to-image synthesis.
\newblock \emph{arXiv preprint arXiv:2306.09341}, 2023{\natexlab{a}}.

\bibitem[Wu et~al.(2023{\natexlab{b}})Wu, Sun, Zhu, Zhao, and Li]{wu2023human}
Wu, X., Sun, K., Zhu, F., Zhao, R., and Li, H.
\newblock Human preference score: Better aligning text-to-image models with human preference.
\newblock In \emph{Proceedings of the IEEE/CVF International Conference on Computer Vision}, pp.\  2096--2105, 2023{\natexlab{b}}.

\bibitem[Wu et~al.(2024)Wu, Hao, Zhang, Sun, Huang, Song, Liu, and Li]{wu2024deep}
Wu, X., Hao, Y., Zhang, M., Sun, K., Huang, Z., Song, G., Liu, Y., and Li, H.
\newblock Deep reward supervisions for tuning text-to-image diffusion models.
\newblock \emph{arXiv preprint arXiv:2405.00760}, 2024.

\bibitem[Xu et~al.(2024{\natexlab{a}})Xu, Liu, Wu, Tong, Li, Ding, Tang, and Dong]{xu2024imagereward}
Xu, J., Liu, X., Wu, Y., Tong, Y., Li, Q., Ding, M., Tang, J., and Dong, Y.
\newblock Imagereward: Learning and evaluating human preferences for text-to-image generation.
\newblock \emph{Advances in Neural Information Processing Systems}, 36, 2024{\natexlab{a}}.

\bibitem[Xu et~al.(2024{\natexlab{b}})Xu, Zhao, Xiao, and Hou]{xu2024ufogen}
Xu, Y., Zhao, Y., Xiao, Z., and Hou, T.
\newblock Ufogen: You forward once large scale text-to-image generation via diffusion gans.
\newblock In \emph{Proceedings of the IEEE/CVF Conference on Computer Vision and Pattern Recognition}, pp.\  8196--8206, 2024{\natexlab{b}}.

\bibitem[Yin et~al.(2024)Yin, Gharbi, Park, Zhang, Shechtman, Durand, and Freeman]{yin2024improved}
Yin, T., Gharbi, M., Park, T., Zhang, R., Shechtman, E., Durand, F., and Freeman, W.~T.
\newblock Improved distribution matching distillation for fast image synthesis.
\newblock \emph{arXiv preprint arXiv:2405.14867}, 2024.

\bibitem[Zhang et~al.(2024)Zhang, Wu, Ren, Xia, Kuang, Xie, Li, Xiao, Huang, Zheng, et~al.]{zhang2024unifl}
Zhang, J., Wu, J., Ren, Y., Xia, X., Kuang, H., Xie, P., Li, J., Xiao, X., Huang, W., Zheng, M., et~al.
\newblock Unifl: Improve stable diffusion via unified feedback learning.
\newblock \emph{arXiv preprint arXiv:2404.05595}, 2024.

\bibitem[Zhao et~al.(2020)Zhao, Liu, Lin, Zhu, and Han]{zhao2020differentiable}
Zhao, S., Liu, Z., Lin, J., Zhu, J.-Y., and Han, S.
\newblock Differentiable augmentation for data-efficient gan training.
\newblock \emph{Advances in neural information processing systems}, 33:\penalty0 7559--7570, 2020.

\end{thebibliography}
\bibliographystyle{icml2025}
.

%%%%%%%%%%%%%%%%%%%%%%%%%%%%%%%%%%%%%%%%%%%%%%%%%%%%%%%%%%%%%%%%%%%%%%%%%%%%%%%
%%%%%%%%%%%%%%%%%%%%%%%%%%%%%%%%%%%%%%%%%%%%%%%%%%%%%%%%%%%%%%%%%%%%%%%%%%%%%%%
% APPENDIX
%%%%%%%%%%%%%%%%%%%%%%%%%%%%%%%%%%%%%%%%%%%%%%%%%%%%%%%%%%%%%%%%%%%%%%%%%%%%%%%
%%%%%%%%%%%%%%%%%%%%%%%%%%%%%%%%%%%%%%%%%%%%%%%%%%%%%%%%%%%%%%%%%%%%%%%%%%%%%%%
\newpage
\appendix
\onecolumn

% You can have as much text here as you want. The main body must be at most $8$ pages long.
% For the final version, one more page can be added.
% If you want, you can use an appendix like this one.  

% The $\mathtt{\backslash onecolumn}$ command above can be kept in place if you prefer a one-column appendix, or can be removed if you prefer a two-column appendix.  Apart from this possible change, the style (font size, spacing, margins, page numbering, etc.) should be kept the same as the main body.
\newpage
\onecolumn
\renewcommand{\thetable}{A\arabic{table}}
\renewcommand{\thefigure}{A\arabic{figure}}
\renewcommand{\theequation}{A\arabic{equation}}

\setcounter{table}{0}
\setcounter{figure}{0}
\setcounter{equation}{0}

\section{Appendix}
\subsection{Gradient Explosion in  DRTune~\cite{wu2024deep}}
\begin{figure}[h]
\vspace{-4mm}
    \centering
    \subfigure[DDIM ]{\includegraphics[width=0.25\linewidth]{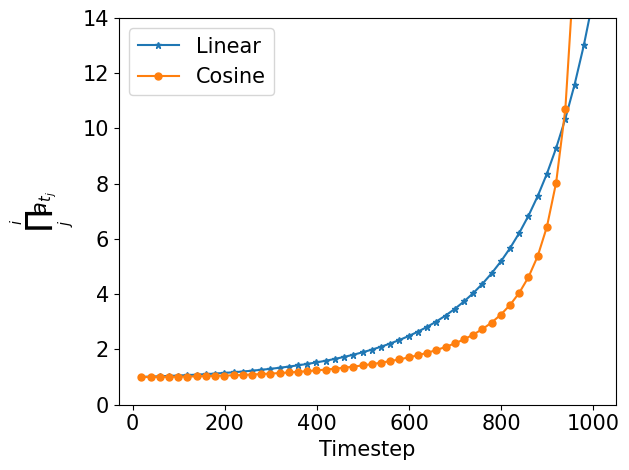}}
    \subfigure[DPMSlover++ ]{\includegraphics[width=0.25\linewidth]{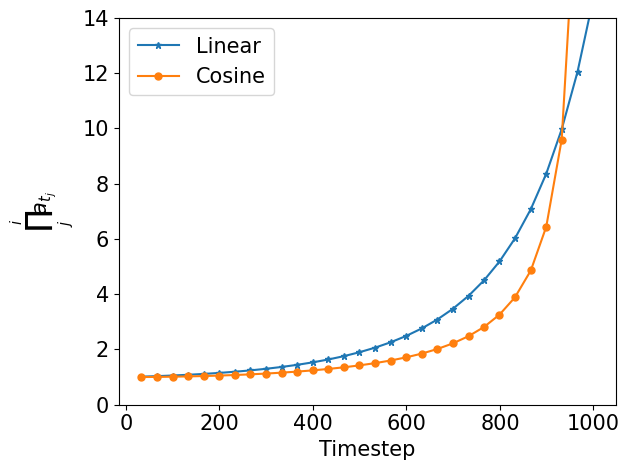}}
    \vspace{-2mm}
    \caption{The curve of gradient scale $\prod_{j=1}^i a_{t_{j}}$ with different samplers.  }
    \vspace{-4mm}
    \label{fig:ge}
\end{figure}

When using DRTune~\cite{wu2024deep} strategy for gradient back-propagation along with the reverse inference path, we can derive the partial differential derivative of the loss function $\mathcal{L}$ on the latent image $\hat{x}_{t_i}$ as the Equation 10, i.e.,
\begin{equation} 
    \begin{split}
        \frac{\partial \mathcal{L}(\hat{x}_0)}{\partial \hat{x}_{t_i}} = \frac{\partial  \mathcal{L}(\hat{x}_0)}{\partial \hat{x}_0} \frac{\partial \hat{x}_0}{\partial \hat{x}_{t_i}}= \frac{\partial  \mathcal{L}(\hat{x}_0)}{\partial \hat{x}_0} \prod_{j=1}^i a_{t_j},
    \end{split}
\end{equation}
where $a_{t_i}$ can be specified by the noise scheduler $\{\alpha_{t_i}, \sigma_{t_i}\}_i$ used in the forward diffusion process asn the samplers used the inference process. Specifically, when using DDIM sampler~\cite{ho2020denoising} during inference, we can specify Equation 3 as,
\begin{equation}
    \begin{split}
        x_{t_{i-1}} = \frac{\alpha_{t_{i-1}}}{\alpha_{t_i}} x_{t_i} + (\sqrt{1-\alpha_{t_{i-1}}^2} - \frac{\alpha_{t_{i-1}} \sqrt{1 - \alpha_{t_{i}}}}{\alpha_{t_{i}}}) \epsilon(x_{t_i}, t_i).
    \end{split}
\end{equation}
Then, we can derive $\prod_{j=0}^i a_{t_j}$ as,
\begin{equation}
     \prod_{j=0}^i a_{t_j} = \prod_{j=0}^i \frac{\alpha_{t_{j-1}}}{\alpha_{j_i}}= \frac{1}{\alpha_{t_i}},
\end{equation}
where $\alpha_{t_i}$ decreases to 0 as $t_i$ increases $T$. Although $\alpha_{T}$ would not be set as $0$ in practice, $\prod_{j=0}^i a_{t_i}$ would also increase rapidly as $i$ grows. Figure~\ref{fig:ge} (a) shows the varying curve with 50 inference steps and linear or cosine noise scheduler, commonly used in practice. We also observe a very similar phenomenon for DPMSlover++~\cite{lu2022dpm2} samplers as shown in Figure~\ref{fig:ge} (b), where the number of inference steps is 30. 

\subsection{Network Architecture for the Discriminator}
The right part of Figure 2 shows the overview of the network architecture for our siamese-network-based discriminator. Here, we provide more details. Following StyleGAN2~\cite{sauer2023stylegan}, we apply the ViT-S network used in DINOv2~\cite{oquab2023dinov2} as the feature network $F$, where five transformer layers equally spaced with the layer numbers [2,5,8,11] are connected to the discriminator heads. The differentiable data augmentation~\cite{zhao2020differentiable}, including the translation padding, the
Cutout square, and the color artifacts with the default, are applied on both the input image and the reference real image with the default settings in~\cite{zhao2020differentiable}. As for the discriminator heads, we use the same settings as that in StyleGAN-T~\cite{sauer2023stylegan}, which consists of convolutions and batch normalization with the output channel as 64. The prediction layer $\eta$ uses a similar network architecture to the discriminator heads with the input as 64.

\vspace{-2mm}
\subsection{Detailed Statistics of the Dataset}
\vspace{-1mm}
In our experiments, to assess ADT's ability to align generated and training data distributions, we collect 170k high-resolution (1K) and high-quality text-image pairs from Midjourney v6~\cite{midjourbey}, which have not been used during the pre-training of the benchmark models, i.e., Stable Diffusion v 1.5\cite{rombach2022high}, Stable Diffusion XL~\cite{podell2023sdxl}, and Stable Diffusion 3 medium~\cite{esser2024scaling}. The detailed numerical statistics can be found in Table~\ref{tab:data}. 

\begin{table}[h]
\centering
\caption{The numerical statistics of our dataset.}\label{tab:data}

\setlength\tabcolsep{3pt}  
\begin{tabular}{ccccc}\hline\hline
         & Instances & Tokens & HPS Score & AES Score \\ \hline
Training & 160k      & 49.30  & 28.92     & 6.634     \\
Test     & 10k       & 49.45  & 28.93     & 6.630   \\
\hline\hline
\end{tabular}
\vspace{-2mm}
\end{table}
% Figure~\ref{fig:data_scores} shows the distribution of HPS scores and AES scores in our dataset. We can find that our training and test data have a similar score distribution, with the minimum HPS score as 28.00 and the minimum AES score as 6.25.

% \begin{figure}[h]
%     \centering
%     \subfloat[HPS Score]{\includegraphics[width=0.5\linewidth]{figs//suppl/density_hps.jpg}}
%     \subfloat[AES Score]{\includegraphics[width=0.5\linewidth]{figs//suppl/density_aes.jpg}}
%     \caption{The histogram of HPS and AES scores for our datasets. }
%     \label{fig:data_scores}
% \end{figure}

% \subsection{Detailed Experimental Settings}

\subsection{Efficiency Analysis}
\begin{figure}[h]
    \centering
    \subfigure[FID]{\includegraphics[width=0.3\linewidth]{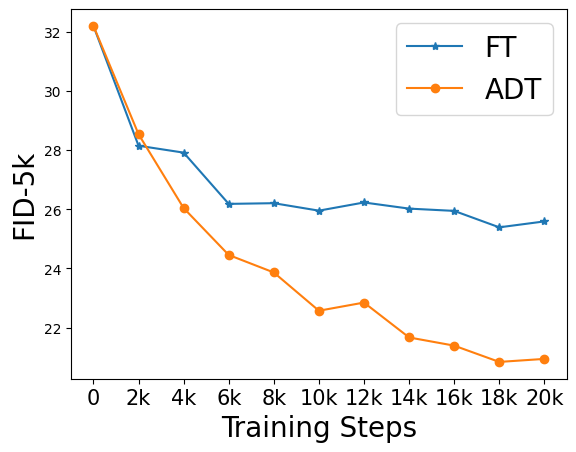}}
    \subfigure[HPS]{\includegraphics[width=0.3\linewidth]{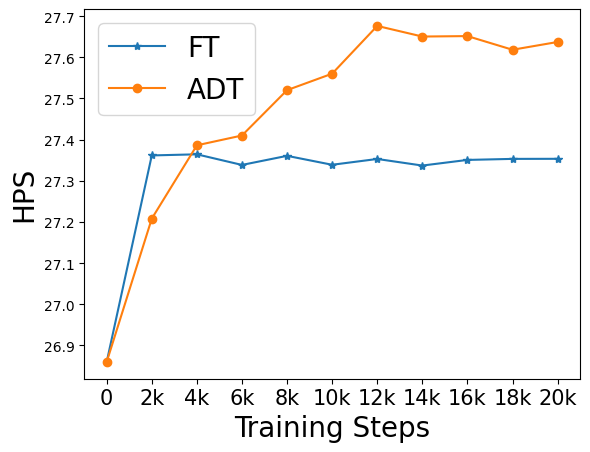}}
    \subfigure[AES]{\includegraphics[width=0.3\linewidth]{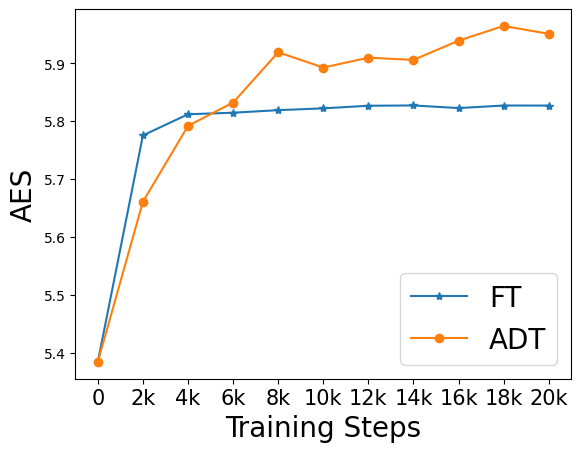}}
    \label{fig:enter-label}
\caption{The performance curve during the optimization for SD15. }\label{fig:steps}
\end{figure}

In our ADT framework, we stimulate the diffusion inference process during optimization and use adversarial supervision on the image space. Intuitively, ADT requires more computing cost compared with the original fine-tuning strategy, where only one model forward is involved during each training step. However, we claim that ADT can outperform the FT strategy with less training steps. Taking SD15 as the example in Figure~\ref{fig:steps}, ADT with 6k training steps has demonstrated better performance compared to the convergence achieved by FT. 

\vspace{-2mm}
\subsection{Experimental Results for the Text Alignment}
\vspace{-1mm}
In our experiments, we have evaluated the performance of ADT with three types of metrics: 1) FID, which measures the divergence of generative distribution and training data distributions. 2) Quality Assessment Model, i.e., HPSv2 Score (\textbf{HPS})~\cite{wu2023human2} and Aesthetics Score (\textbf{AES})~\cite{Christoph2022aes}, which assess the image quality with the assessment models trained on the human preference feedback data. 3) Human-level evaluation of the image quality and text alignment. Here, we further verify the text alignment of the models trained by the ADT framework with the CLIP Score as the metric. Specifically, 5k prompts are randomly selected from our test data for generating images. Table~\ref{tab:clipscore} shows the results with different settings. We can find that our ADT framework can also increase the capability of text alignment in terms of the CLIP score. However, we claim that the CLIP score may not be a good metric when the prompt is too long.

\begin{table}[h]
\centering
\caption{The evaluation of the text alignment with the CLIP score.}\label{tab:clipscore}

\begin{tabular}{cccc} \hline\hline
            & Base   & FT     & ADT    \\\hline
SD15+DDIM50 & 0.3324 & 0.3339 & \textbf{0.3346} \\
SD15+DPMS30 & 0.3313 & 0.3330 & \textbf{0.3340} \\
SDXL+DDIM50 & 0.3677 & 0.3708 & \textbf{0.3759} \\
SDXL+DPMS30 & 0.3712 & 0.3749 & \textbf{0.3758} \\
SD3         & 0.3547 & 0.3616 & \textbf{0.3641}  \\
\hline\hline
\end{tabular}
\vspace{-2mm}
\end{table}

\subsection{How to Choose $\lambda$  }
The hyper-parameter $\lambda$ in Equation 8 aims to balance the adversarial loss and the original diffusion loss. The ablation analysis in Section 5.3 has demonstrated that the original diffusion loss is critical to avoiding knowledge forgetting and discriminator hacking. Here, we further conduct the sensitive analysis for $\lambda$. Taking SD15 as an example, where the DDIM sampler is used with 50 inference steps, we vary the $\lambda$ among [0.1, 0.3, 0.5,0.7, 1.0] and summarize the performances in Table~\ref{tab:lambda}. We can find that our ADT framework outperforms FT with $\lambda>0$ on all three metrics, which demonstrates the superiority and robustness of ADT. In addition, ADT achieves the best scores in FID-5k and AES score when $\lambda =0.5$, which is used in most of our experiments.

\begin{table}[h]
\centering
\caption{The sensitive analysis of the hyper-parameter $\lambda$.}\label{tab:lambda}

\begin{tabular}{cccc} \hline\hline
              & FID-5k & HPS Score & AES Score \\ \hline
FT            & 23.08  & 27.37     & 5.821     \\ 
$\lambda=0.0$   & 24.07  & 27.18    & 5.873     \\ \hline
$\lambda=0.1$ & 21.11  & 27.53    & 5.924     \\
$\lambda=0.3$ & 20.35  & 27.64     & 5.938     \\
$\lambda=0.5$ & \textbf{19.84}  & 27.59     & \textbf{5.952}     \\
$\lambda=0.7$ & 20.03  & 27.65     & 5.924     \\
$\lambda=1.0$ & 19.93  & \textbf{27.66}     & 5.917     \\
\hline\hline
\end{tabular}
\vspace{-2mm}
\end{table}

% \subsection{How to Choose the Paramter $K$  }

\subsection{More discussion about the Feature Network}
In our experiments, the small version of DINOv2 is used for the backbone of the discriminator with 21M parameters. Here we further verify the performance of ADT with a larger feature network $F$, i.e., the large version of DINOv2 with 300M parameters, where five transformer layers with the layer numbers [5, 10, 14, 19, 23] are connected to the discriminator heads. Table~\ref{tab:dino} shows the results of the SD15 model with DDIM samplers and 50 inference steps. Interestingly, the larger pre-trained feature network, which is equipped with more powerful expression ability, has not brought significant improvement in our experiments, which needs more discussion in the future. 
\begin{table}[h]
\centering
\caption{The performance with large feature netwotk.}\label{tab:dino}
\begin{tabular}{cccc} \hline\hline
           & FID   & HPS Score & AES Score \\ \hline
DINO-Small (21M) & 19.84 & 27.59     & 5.952     \\
ADT with DINO-Large (300M) & 20.27 & 27.61     &   5.920       \\
\hline\hline
\end{tabular}
\vspace{-2mm}
\end{table}

% \subsection{More Qualitative Results}
% Here, we show more qualitative results for all benchmark models for clear comparisons.
\begin{figure}
    \centering
    \includegraphics[width=1\linewidth]{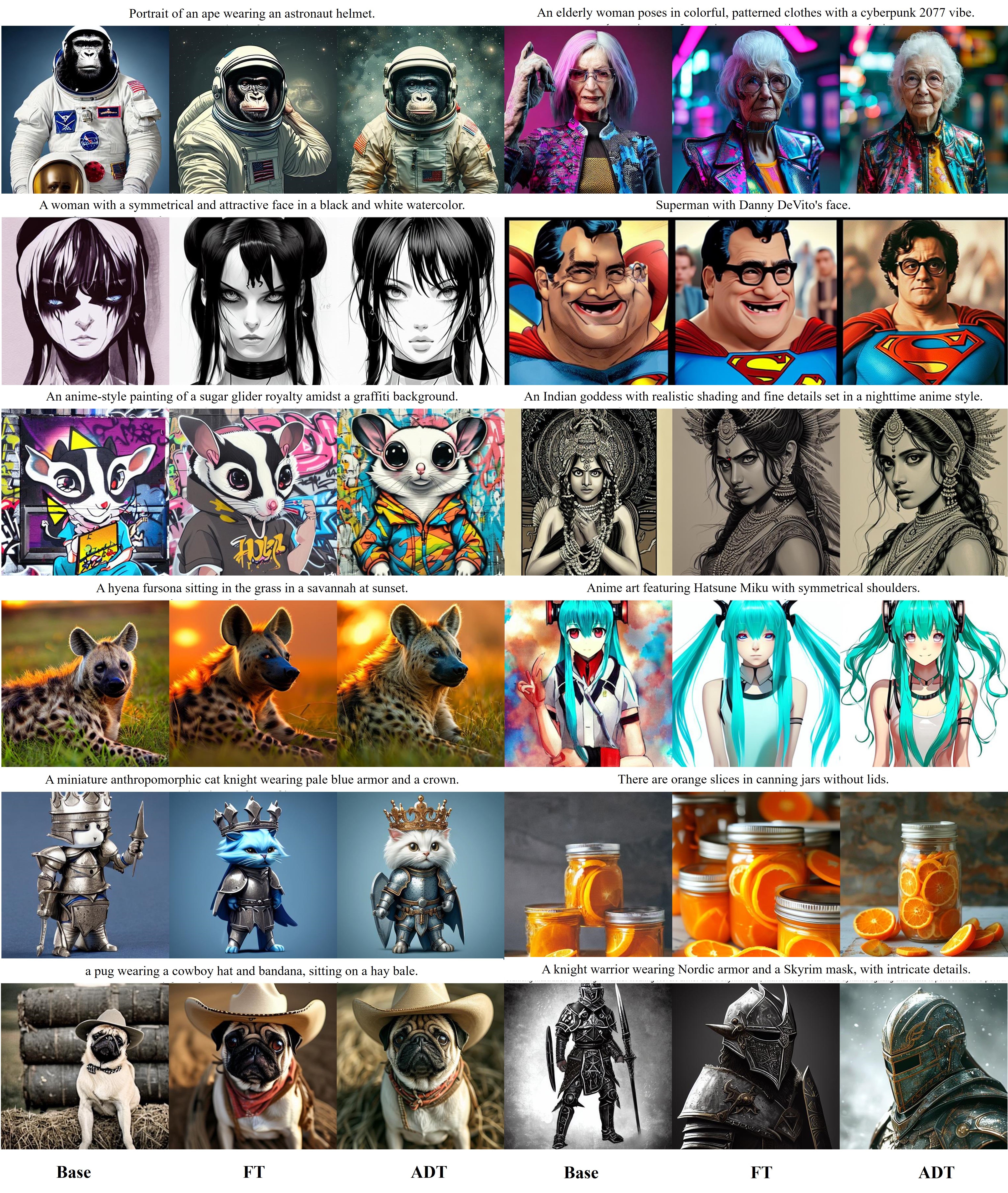}
    \caption{More qualitative results for SD15 with the DDIM sampler and 50 inference steps. Prompts are listed above the picture.}
    \label{fig:sd15_ddim_case}
\end{figure}

\begin{figure}
    \centering
    \includegraphics[width=1\linewidth]{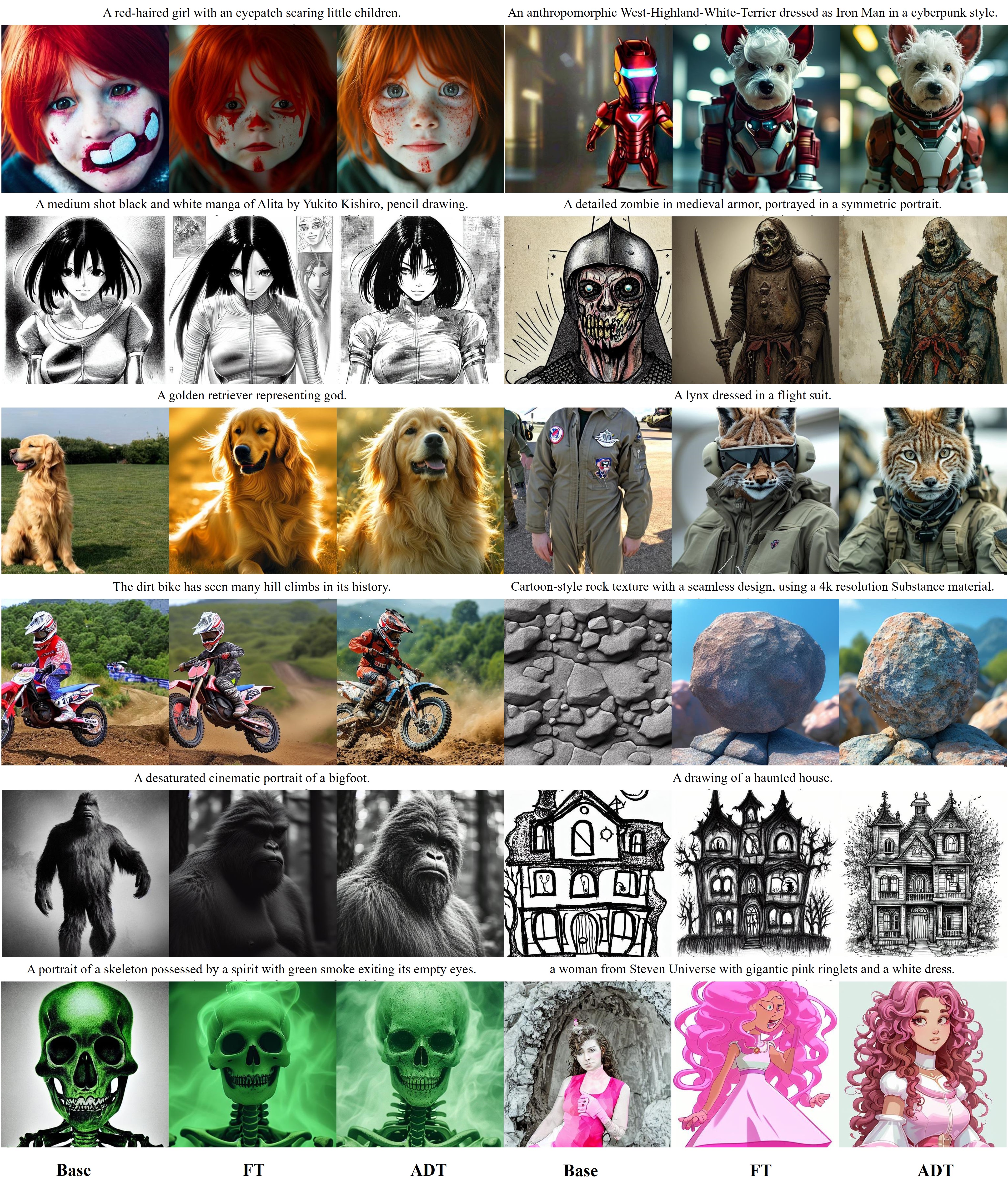}
    \caption{More qualitative results for SD15 with the DPMSlover++ sampler and 30 inference steps. Prompts are listed above the picture.}
    \label{fig:sd15_dpm_case}
\end{figure}

\begin{figure}
    \centering
    \includegraphics[width=0.80\linewidth]{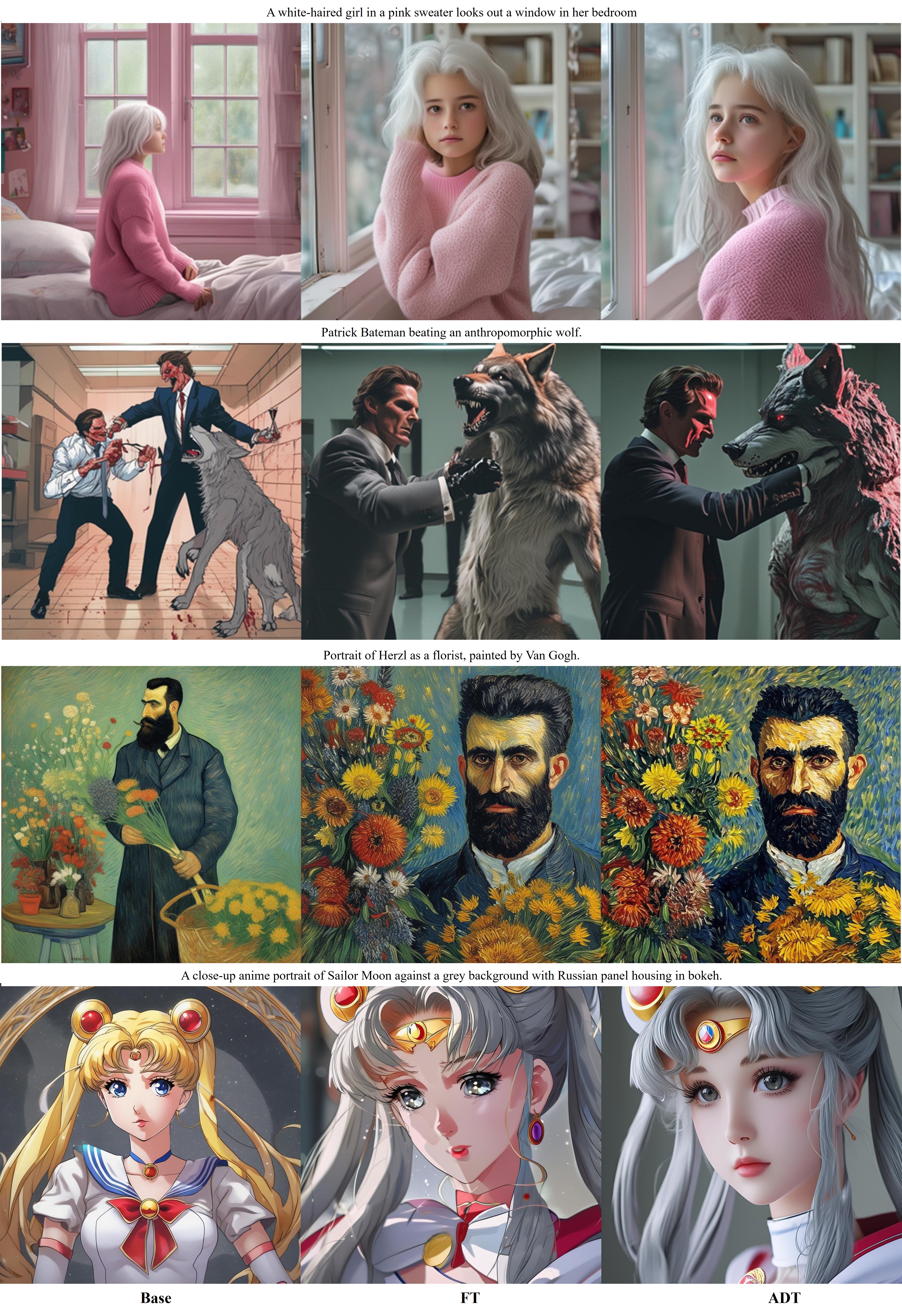}
    \caption{More qualitative results for SDXL with the DDIM sampler and 50 inference steps. Prompts are listed above the picture.}
    \label{fig:sd3}
\end{figure}

\begin{figure}
    \centering
    \includegraphics[width=0.80\linewidth]{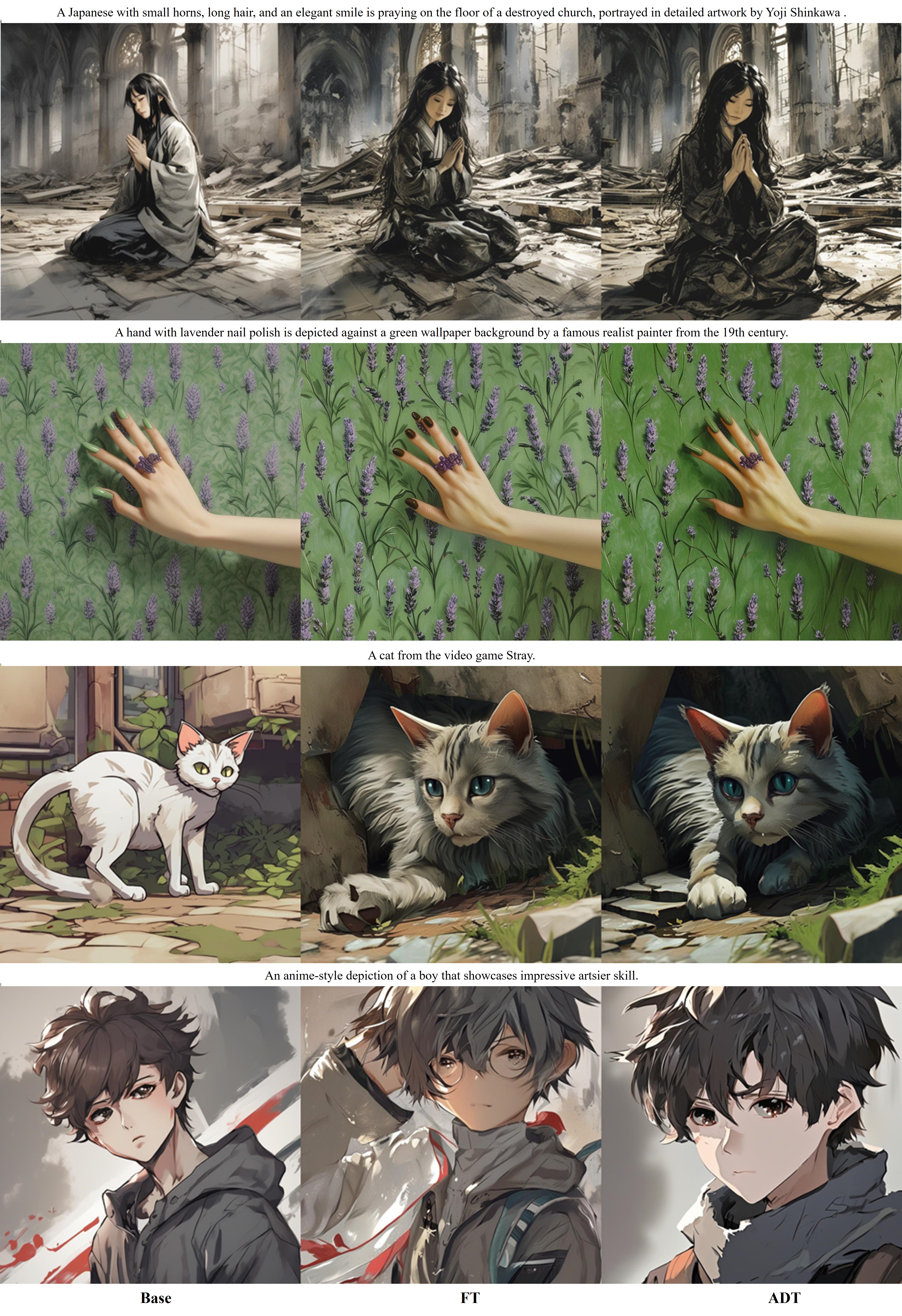}
    \caption{More qualitative results for SDXL with the DDIM sampler and 50 inference steps. Prompts are listed above the picture.}
    \label{fig:sd3}
\end{figure}

\begin{figure}
    \centering
    \includegraphics[width=0.80\linewidth]{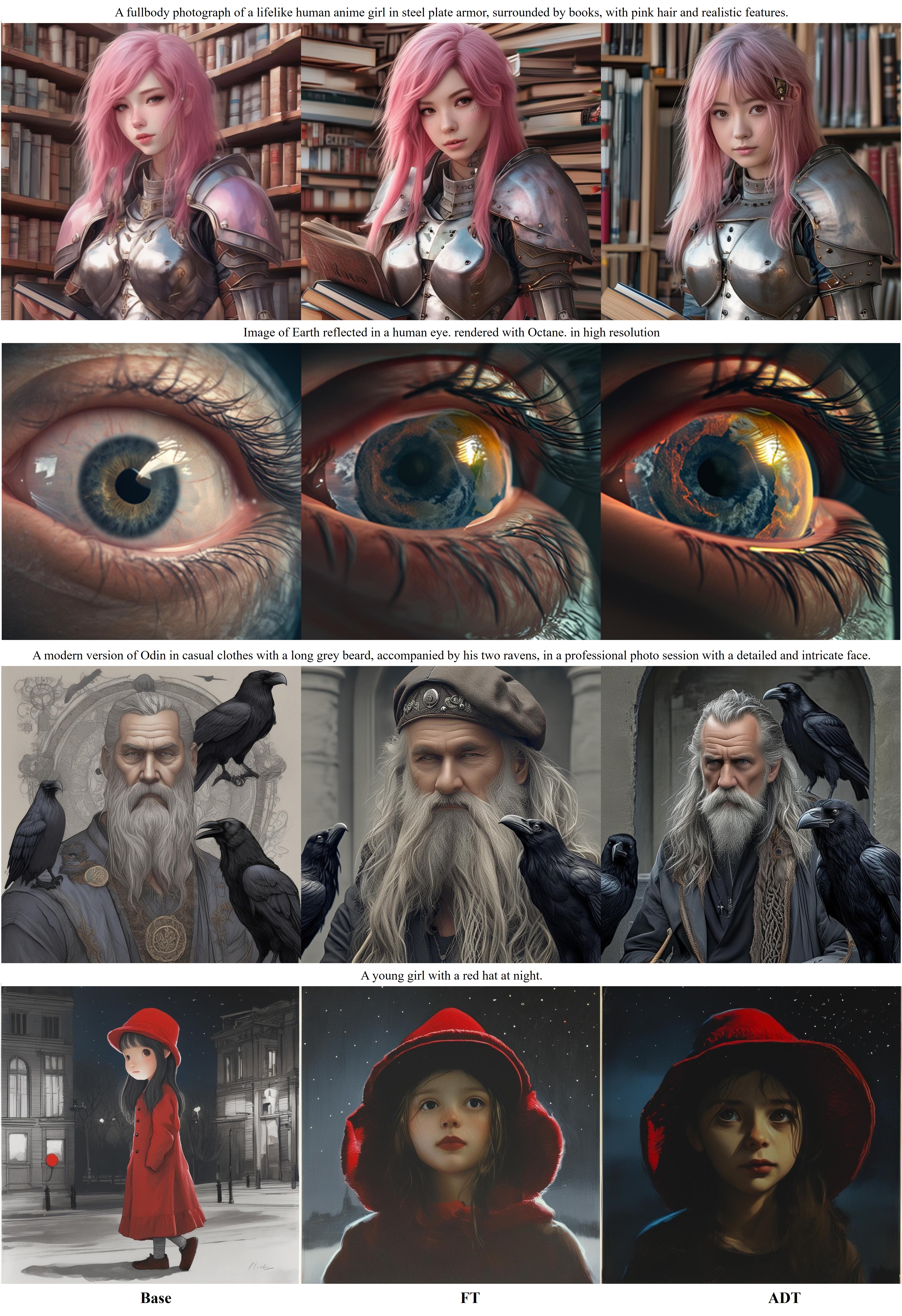}
    \caption{More qualitative results for SDXL with the DPMSlover++ sampler and 30 inference steps. Prompts are listed above the picture.}
    \label{fig:sd3}
\end{figure}

\begin{figure}
    \centering
    \includegraphics[width=0.80\linewidth]{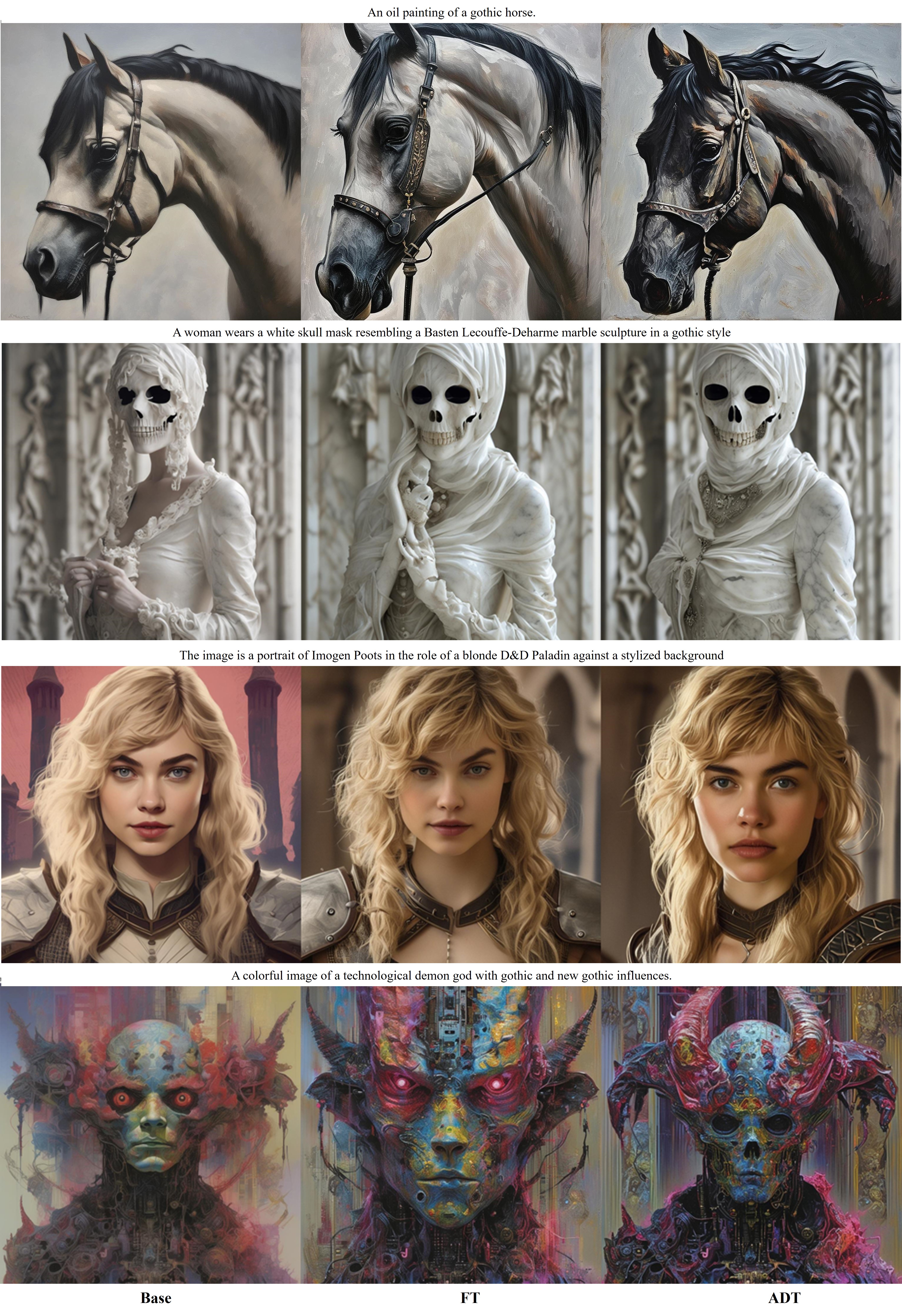}
    \caption{More qualitative results for SDXL with the DPMSlover++ sampler and 30 inference steps. Prompts are listed above the picture.}
    \label{fig:sd3}
\end{figure}

\begin{figure}
    \centering
    \includegraphics[width=0.80\linewidth]{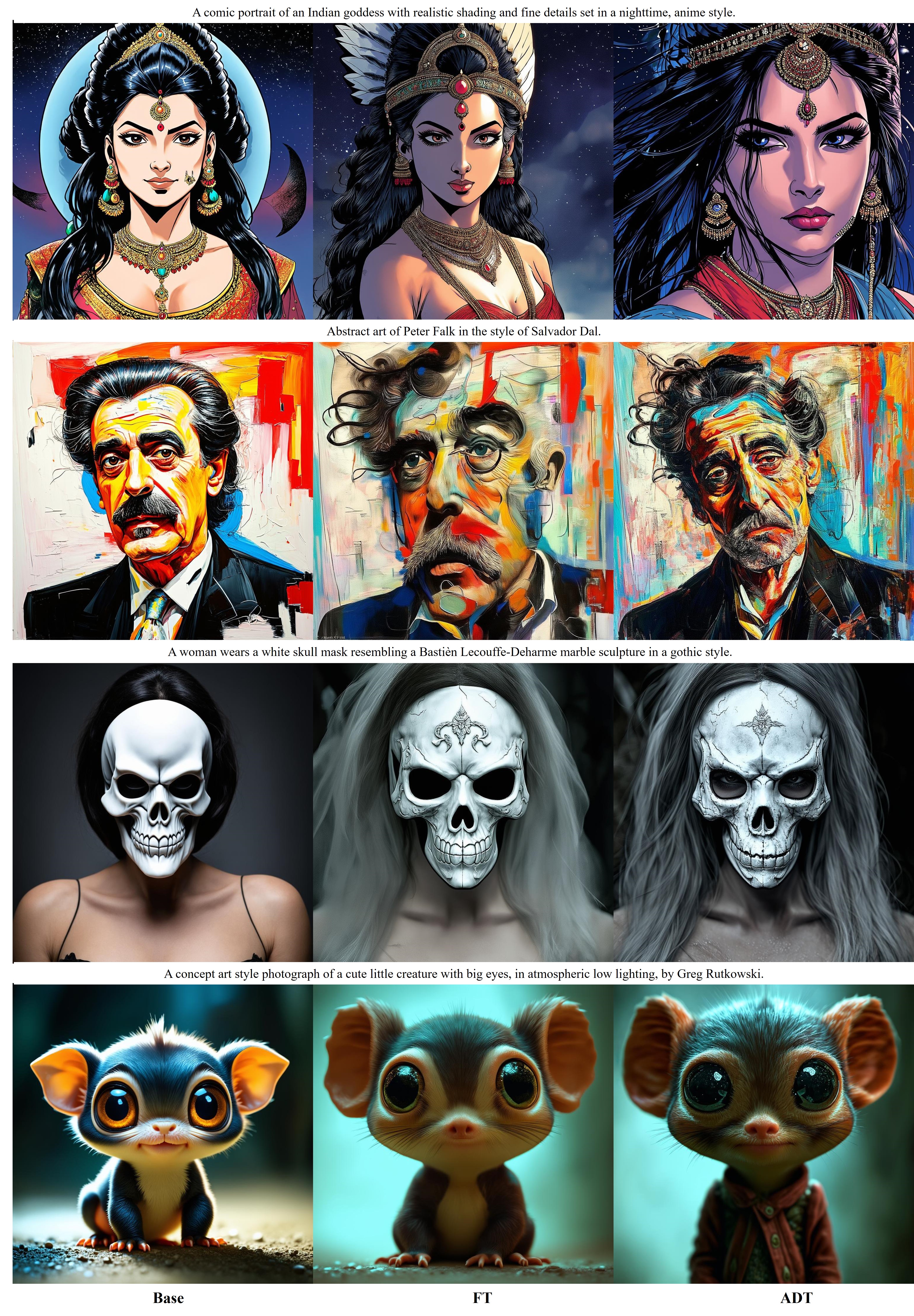}
    \caption{More qualitative results for SD3. Prompts are listed above the picture.}
    \label{fig:sd3}
\end{figure}

\begin{figure}
    \centering
    \includegraphics[width=0.80\linewidth]{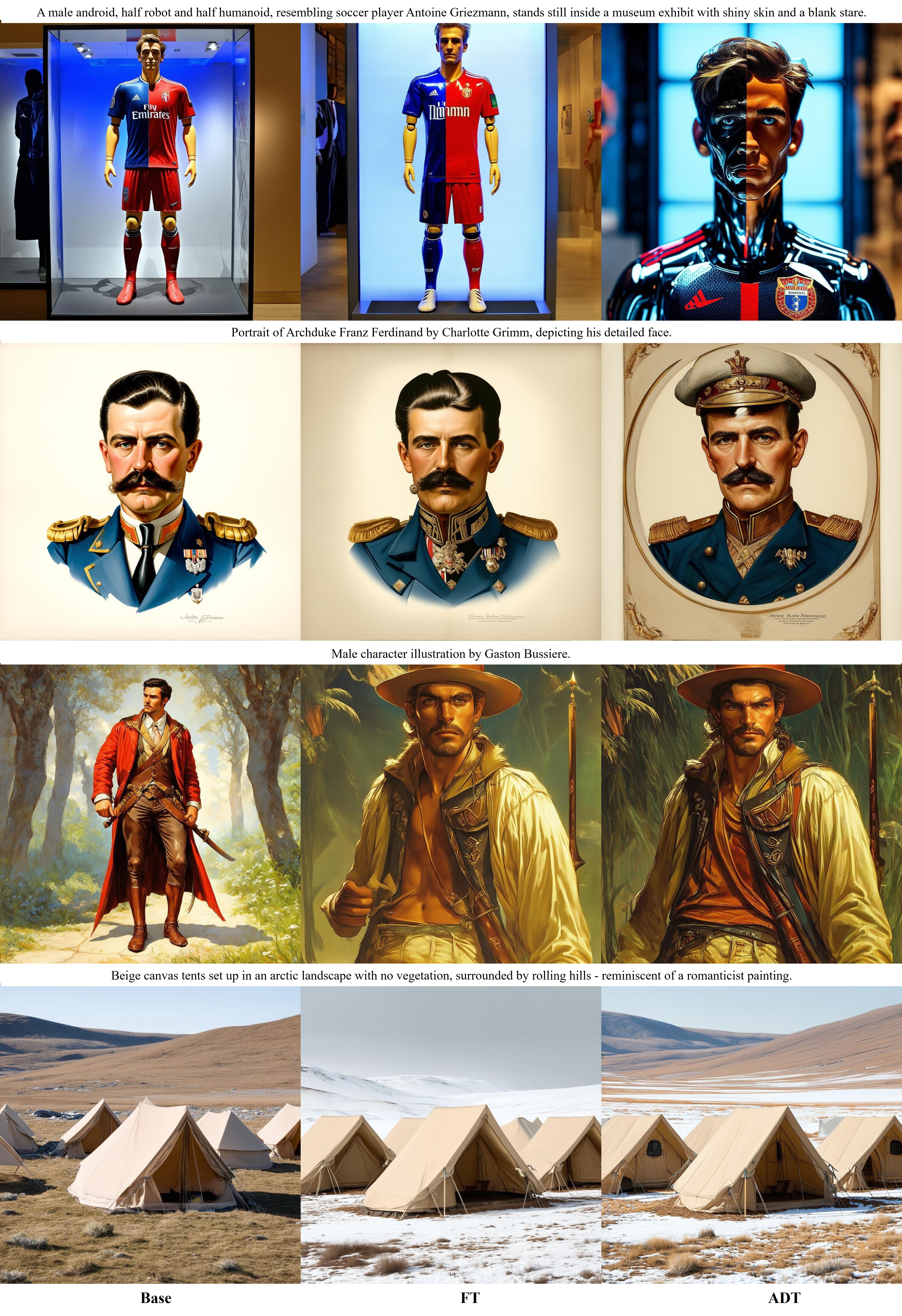}
    \caption{More qualitative results for SD3. Prompts are listed above the picture.}
    \label{fig:sd3}
\end{figure}
%%%%%%%%%%%%%%%%%%%%%%%%%%%%%%%%%%%%%%%%%%%%%%%%%%%%%%%%%%%%%%%%%%%%%%%%%%%%%%%
%%%%%%%%%%%%%%%%%%%%%%%%%%%%%%%%%%%%%%%%%%%%%%%%%%%%%%%%%%%%%%%%%%%%%%%%%%%%%%%

\end{document}